\documentclass{article}
\usepackage{authblk}
\usepackage[utf8]{inputenc}
\usepackage{main}
\usepackage{microtype}
\usepackage{subcaption}
\usepackage{graphicx}
\usepackage{times}
\usepackage{latexsym}
\usepackage{amsmath}
\usepackage{float}
\usepackage{footnote}
\usepackage{enumitem}
\usepackage{bm}
\usepackage{arydshln}
\usepackage{booktabs}
\usepackage{multicol}
\usepackage{multirow}
\usepackage{color}
\usepackage{xcolor}     
\usepackage{colortbl}
\usepackage{bbding}
\usepackage{makecell}
\usepackage{mathtools}
\usepackage{imakeidx}
\usepackage{longtable}
\usepackage{wrapfig}
\usepackage{rotating}
\makeindex
\usepackage{arydshln}
\usepackage{lipsum}
\usepackage{natbib}
\usepackage[toc]{multitoc}
\usepackage[edges]{forest}
\usepackage[normalem]{ulem}
\definecolor{mydarkblue}{rgb}{0,0.08,0.45}
\usepackage[colorlinks=true,linkcolor=mydarkblue,citecolor=mydarkblue,filecolor=mydarkblue,urlcolor=mydarkblue]{hyperref}
\usepackage{CJKutf8}
\usepackage{awesomebox} 
\usepackage{bbding}
\usepackage[most]{tcolorbox}
\usepackage{booktabs}
\usepackage{geometry}
\definecolor{wkblue}{rgb}{0.2, 0.3, 0.6}
\definecolor{meta-color}{rgb}{0.5, 0.5, 0.5}
\usepackage{amsmath}
\usepackage{enumitem}
\usepackage{lscape} 
\usepackage{booktabs}
\usepackage{algorithm}      
\usepackage{algorithmicx}
\usepackage{algpseudocode}
\usepackage{tabularx,booktabs}
\usepackage{makecell}

\usepackage{amssymb}
\usepackage{amsfonts}

\usepackage[tikz]{bclogo}
\usepackage[framemethod=tikz]{mdframed}
\definecolor{bgblue}{RGB}{245,243,253}
\definecolor{ttblue}{RGB}{91,194,224}

\usepackage{fancyhdr} 
\usepackage{blindtext} 
\usepackage{makecell}

\usepackage{tikz}
\usepackage[edges]{forest}
\usepackage{microtype}
\usepackage{hyperref}
\usepackage{url}
\usepackage{booktabs}
\usepackage{bbding}
\usepackage{multirow}
\usepackage{booktabs}
\usepackage{float}
\usepackage{makecell}
\usepackage{times}
\usepackage{latexsym}
\usepackage{url}
\usepackage{forest}
\usepackage{setspace}
\usepackage{placeins}
\usepackage{booktabs}
\usepackage{multirow}
\usepackage{tabularx}
\usepackage{tabulary}
\usepackage{graphicx}
\usepackage{adjustbox}
\usepackage{colortbl}
\usepackage{xcolor}
\usepackage{enumitem}
\usepackage{amsfonts}
\usepackage{bbding}
\usepackage{wrapfig}
\usepackage{amsmath}
\usepackage{hyperref}
\usepackage{pifont}
\usetikzlibrary{trees,positioning,shapes,shadows,arrows.meta}
\usepackage{chemformula}

\usepackage[edges]{forest}
\usepackage[most]{tcolorbox}
\usepackage{lipsum} 
\usepackage{fontawesome5}  

\newtcolorbox{QABox}[1][]{%
  enhanced,
  breakable,
  sharp corners,
  colback=blue!2!white,
  colframe=orange!10!white,
  coltitle=black,
  title={\faLightbulb[regular]~Hands-on Guidelines: #1},
  fonttitle=\bfseries\large,
  boxrule=0.8pt,
  top=6pt,
  bottom=6pt,
  left=8pt,
  right=8pt,
  width=\textwidth,
}

\newcommand{\Q}[1]{\textbf{\faQuestionCircle~Q:}~#1}
\newcommand{\A}[1]{\textbf{\faCheckCircle~A:}~#1}
\newcommand{\eg}{\textit{e}.\textit{g}.}
\newcommand{\ie}{\textit{i}.\textit{e}.}
\newcommand{\TTS}{\textit{TTS }}
\definecolor{hidden-draw}{RGB}{195,82,66}
\definecolor{hidden-blue}{RGB}{194,232,247}
\definecolor{hidden-orange}{RGB}{195,82,66}
\definecolor{hidden-yellow}{RGB}{242,244,193}
\definecolor{tree-level-1}{RGB}{195,82,66}
\definecolor{tree-level-2}{RGB}{195,82,66}
\definecolor{tree-level-3}{RGB}{195,82,66}
\definecolor{tree-leaf}{RGB}{195,82,66}
\newcommand{\cmark}{\textcolor{green!70!black}{\ding{51}}}

\newcommand{\xmark}{\textcolor{red}{\ding{55}}}

\usepackage{microtype}
\usepackage{bbding}
\definecolor{darkblue}{rgb}{0, 0, 0.5}
\hypersetup{colorlinks=true, citecolor=darkblue, linkcolor=darkblue, urlcolor=darkblue}

\usepackage{lineno}

\newcommand*\samethanks[1][\value{footnote}]{\footnotemark[#1]}
\title{A Survey on Test-Time Scaling in Large Language Models: \\ What, How, Where, and How Well}

\author{Qiyuan Zhang$^{1}$\thanks{~~Core contribution}\hspace{0.5em}, Fuyuan Lyu$^{2}$\samethanks\hspace{0.5em}, 
Zexu Sun$^{3}$\thanks{~~Significant contribution}\hspace{0.5em}, Lei Wang$^{5}$\samethanks\hspace{0.5em}, Weixu Zhang$^{2}$\samethanks\hspace{0.5em}, Wenyue Hua$^{8}$\samethanks\hspace{0.5em}, Haolun Wu$^{2,7}$\samethanks\hspace{0.5em}, Zhihan Guo$^{4}$\samethanks\hspace{0.5em}, 
Yufei Wang$^{6}$\thanks{~~Taxonomy designer}\hspace{0.5em}, Niklas Muennighoff$^{7}$, Irwin King$^{4}$, Xue Liu$^{2}$, Chen Ma$^{1}$\\

$^{1}$City University of Hong Kong,
$^{2}$McGill University \& MILA,
$^{3}$Gaoling School of Artificial Intelligence, Renmin University of China,
$^{4}$Chinese University of Hong Kong,
$^{5}$Salesforce AI Research,   
$^{6}$Macquarie University,
$^{7}$Stanford University,
$^{8}$University of California, Santa Barbara\\
\texttt{qzhang732-c@my.cityu.edu.hk}, 
\texttt{fuyuan.lyu@mail.mcgill.ca} \\
\texttt{\quad}\\
\textbf{Page:} \texttt{\url{https://testtimescaling.github.io/}}}
\begin{document}
\maketitle

\begin{abstract}
As enthusiasm for scaling computation (data and parameters) in the pretraining era gradually diminished, test-time scaling (\textit{TTS})\textemdash{}also referred to as ``test-time computing''\textemdash{}has emerged as a prominent research focus. Recent studies demonstrate that \textit{TTS} can further elicit the problem-solving capabilities of large language models (LLMs), enabling significant breakthroughs not only in specialized reasoning tasks, such as mathematics and coding, but also in general tasks like open-ended Q\&A.
However, despite the explosion of recent efforts in this area,
there remains an urgent need for a comprehensive survey offering systemic understanding.
To fill this gap, we propose a unified, hierarchical framework structured along four core dimensions of \textit{TTS} research: \textbf{\textit{what to scale}}, \textbf{\textit{how to scale}}, \textbf{\textit{where to scale}}, and \textbf{\textit{how well to scale}}. 
Building upon this taxonomy, we conduct an extensive review of methods, application scenarios, and assessment aspects, and present an organized decomposition that highlights the unique contributions of individual techniques within the broader \textit{TTS} landscape. From this analysis, we distill the major developmental trajectories of \textit{TTS} to date and offer hands-on guidelines for practical deployment. Furthermore, we identify several open challenges and offer insights into promising future directions, including further scaling, clarifying the functional essence of techniques, generalizing to more tasks, and more attributions.
Our repository is available on \url{https://github.com/testtimescaling/testtimescaling.github.io/}.
\end{abstract}

\section{Introduction}
\label{sec:intro}
Large language models (LLMs)~\citep{brown2020languagemodelsfewshotlearners,openai-gpt4} have emerged in recent years as a transformative milestone toward artificial general intelligence (AGI)~\citep{Goertzel2014148,bubeck2023sparksartificialgeneralintelligence}.
These models remarkably learn general intelligence by \emph{training-time scaling}, where the models ingest more data and parameters~\citep{kaplan2020scalinglawsneurallanguage, hoffmann2022trainingcomputeoptimallargelanguage}. 
However, the progress of pretraining scaling has gradually slowed due to its resource-intensive nature and the bounded availability of human data, prompting researchers to shift their focus toward how to fully elicit the intelligence encoded in LLMs at test time to maximize their real-world effectiveness~\citep{wei2022chain, ouyang2022training,li2024chainthoughtempowerstransformers}?

\begin{figure}[!htbp]
    \centering
    \includegraphics[width=0.98\linewidth]{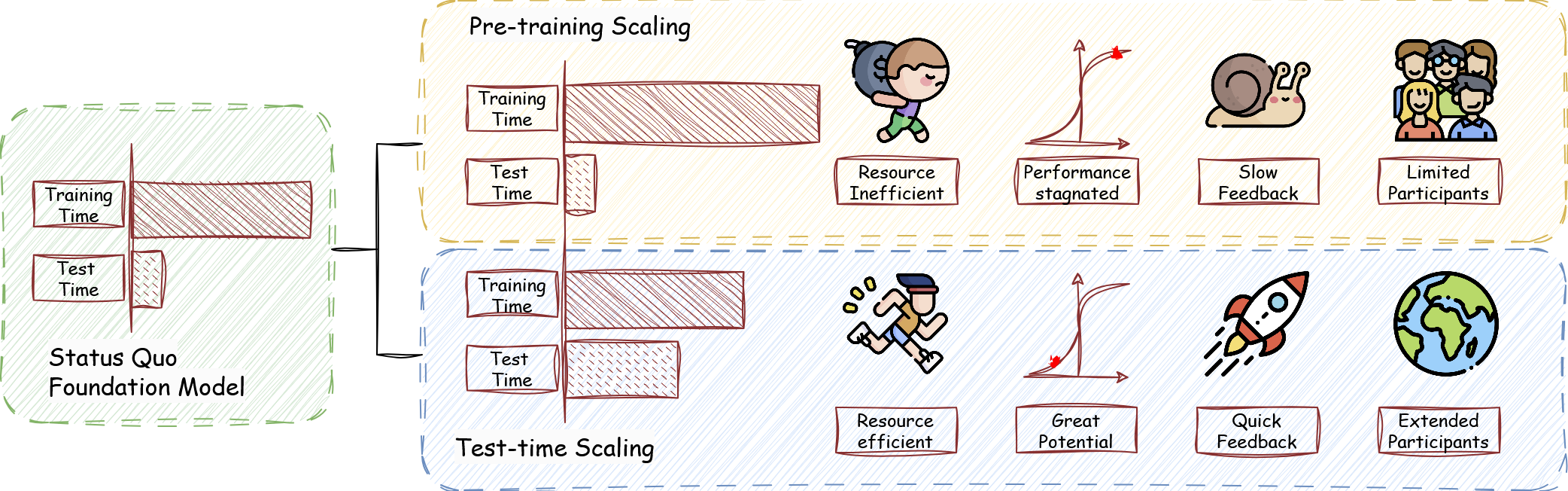}
    \caption{Comparison of Scaling Paradigms in Pre-training and Test-time Phases.}
    \label{fig:intro}
\end{figure}

Human cognition may suggest a clue. When faced with complex problems, people tend to engage in deeper, more deliberate thinking, often producing better outcomes~\citep{kahneman2011thinking, daniel2003maps, Evans1984-EVAHAA}. 
Inspired by this principle, recent research~\citep{wei2022chain, wang2023selfconsistency} has introduced methods that \emph{allocate additional computation during inference} to boost task performance.
Notably, some studies~\citep{brown2024large, wu2024scaling} observe patterns akin to scaling laws: increasing inference-time compute yields consistent performance improvements.
This family of methods, referred to as \textbf{test-time scaling} (\textbf{\textit{TTS}}), progressively elicits the model’s intelligence in the test-time, as depicted in Figure~\ref{fig:intro}.
The remarkable successes of reasoning models, such as \textit{o1}~\citep{openai-o1} and \textit{R1}~\citep{deepseek-r1},
have further amplified interest in \textit{TTS}, highlighting its potential as a key driver of LLM reasoning and utility.
However, despite this surge in research activity, the field currently lacks a unified and systematic framework to synthesize insights, compare techniques, or identify consistent trends in \TTS. To address this gap, we present a comprehensive survey of \textit{TTS}, offering a hierarchical and extensible framework to analyze methods, map research efforts, and guide future progress.

To the best of our knowledge, this is the first survey to comprehensively examine \textit{TTS} \emph{across multiple orthogonal dimensions}, offering a structured perspective for both theoretical inquiry and practical deployment.
Our framework dissects \textit{TTS} into four key dimensions: \textbf{\textit{what to scale}}, \textbf{\textit{how to scale}}, \textbf{\textit{where to scale}}, and \textbf{\textit{how well to scale}}. Our work emphasizes a fine-grained, decomposition-based understanding of \textit{TTS}. 
While prior efforts have examined \textit{TTS} from specific lenses—such as input modification and output verification~\citep{snell2024scaling}, or through the lens of \textit{System 2 AI} and \textit{Long Chain-of-Thought} (CoT)~\citep{li2025system, ji2025test, chen2025reasoningerasurveylong}—these works are structured around a timeline, tracing the evolution of techniques over time.
We analyze the full pipeline, from scaling formulations and algorithmic mechanisms to task domains and performance dimensions. We provide a structured foundation that allows future research to be seamlessly integrated into our taxonomy, making it easier to understand their contributions. 
Specifically, \textit{what to scale} in Sec.~\ref{sec:what2scale} is about what to be scaled at the inference stage. \textit{How to scale} in Sec.~\ref{sec:how2scale} depicts how they are implemented. We categorize various techniques, recognizing that a single work may involve multiple techniques; 
\textit{Where to scale} in Sec.~\ref{sec:where2scale} covers the tasks and datasets where these techniques are applied. Finally, \textit{how well to scale} inSec.~\ref{sec:howwell2scale} refers to evaluating the different attributions of \textit{TTS} methods. 
We further provide fine-grained subcategories under each axis and systematically map representative works to highlight their contributions and trade-offs (Sec.~\ref{sec:organizationandtrends}).
From this structured analysis, we extract major trends in \textit{TTS} development and offer \emph{hands-on guidance} (Sec.~\ref{sec:handon}) for real-world deployment. Grounded in our multi-dimensional taxonomy, we also identify persistent challenges and promising research directions (Sec.~\ref{sec:challenges}): advancing test-time scalability, clarifying the fundamental essence of the effectiveness of different techniques in \textit{TTS}, broadening generalization to a wider range of downstream tasks, and optimizing \textit{TTS} methods along additional dimensions such as efficiency.

\paragraph{Our contributions are threefold:}
\begin{enumerate}
    \item \textbf{A Unified, Multi-Dimensional Taxonomy.} We propose a four-axis taxonomy—\textit{what to scale}, \textit{how to scale}, \textit{where to scale}, and \textit{how well to scale}—that supports structured classification, comparison, and extensibility for \textit{TTS} methods.
    \item \textbf{Systematical Literature Organization and Pragmatic Analysis.} 
    Using our taxonomy, we survey the \textit{TTS} landscape, analyze representative methods, and present guidelines for research application and deployment.
    \item \textbf{Challenges, Insights, and Forward Directions.} Building on our organized perspective, we uncover critical challenges—ranging from advancing scaling to clarifying essence—and outline promising research directions that could shape future progress. Our unified framework facilitates the mapping of these open questions to concrete dimensions of \textit{TTS}, enabling more targeted and impactful advancements.
\end{enumerate}

We plan to continuously update our taxonomy to reflect ongoing progress and provide an evolving foundation for organizing future developments in \textit{TTS} research.
\tikzstyle{my-box}= [
    rectangle,
    draw=hidden-draw,
    rounded corners,
    text opacity=1,
    minimum height=1.5em,
    minimum width=5em,
    inner sep=2pt,
    align=center,
    fill opacity=.5,
]
\tikzstyle{leaf}=[
my-box, 
minimum height=1.5em,
fill=yellow!27, 
text=black,
align=left,
font=\scriptsize,
inner xsep=5pt,
inner ysep=4pt,
align=left,
text width=45em,
]
\tikzstyle{leaf2}=[
my-box, 
minimum height=1.5em,
fill=purple!22, 
text=black,
align=left,
font=\scriptsize,
inner xsep=5pt,
inner ysep=4pt,
]
\tikzstyle{leaf3}=[
my-box, 
minimum height=1.5em,
fill=hidden-blue!57, 
text=black,
align=left,
font=\scriptsize,
inner xsep=5pt,
inner ysep=4pt,
]
\tikzstyle{leaf4}=[
my-box, 
minimum height=1.5em,
fill=green!17, 
text=black,
align=left,
font=\scriptsize,
inner xsep=5pt,
inner ysep=4pt,
]
\begin{figure}[!htbp]
\vspace{-2.5em}
\centering
\scalebox{0.7}{
  \begin{forest}
    forked edges,
    for tree={
      grow=east,
      reversed=true,
      anchor=base west,
      parent anchor=east,
      child anchor=west,
      base=left,
      font=\small,
      rectangle,
      draw=hidden-orange,
      rounded corners,
      align=left,
      minimum width=4em,
      edge+={darkgray, line width=1pt},
      s sep=3pt,
      inner xsep=2pt,
      inner ysep=3pt,
      ver/.style={rotate=90, child anchor=north, parent anchor=south, anchor=center},
    },
    where level=1{text width=4.5em,font=\scriptsize,}{},
    where level=2{text width=5.4em,font=\scriptsize,}{},
    where level=3{text width=6.4em,font=\scriptsize,}{},
    where level=4{text width=6.4em,font=\scriptsize,}{},
[
    \textbf{Test-time Scaling}, ver
    [
        \textbf{What to Scale}\\ (\S \ref{sec:what2scale})
        [
            \textbf{Parallel Scaling} \\ (\S \ref{subsec:parallelsclaing})
            [
                Self-Consistency~\citep{brown2024large, irvine2023rewarding, song2024good, snell2024scaling, wang2023selfconsistency, nguyen2024consistent} \\ 
                \citep{chen2024are, wu2025lessunderstandingchainofthoughtlength}{,} 
                Multi-Agents~\citep{jiang2023llm}{,}
                PlanSearch~\citep{wang2024planningnaturallanguageimproves}{,}
                CCE~\citep{zhang2025crowd}, leaf, text width=40em
            ]
        ]
        [
            \textbf{Sequential Scaling}\\ (\S \ref{subsec:sequentialsclaing})
            [
                Self-Refine~\citep{madaan2023selfrefine, chen2024teaching, gou2024critic, zhang2024smalllanguagemodelsneed}{,}
                Sequential Revision~\citep{lee2025evolvingdeeperllmthinking}{,} ReAct\\
                \citep{yao2023react}{,}
                Budget-aware~\citep{kimi-k1.5, muennighoff2025s1, han2025tokenbudgetawarellmreasoning}{, }RecurrentBlock~\citep{geiping2025scalingtesttimecomputelatent}{,} STaR\\
                \citep{yuan2023scaling, singh2024beyond}{,} 
                Meta-STaR~\citep{xiang20252reasoningllmslearning}{,}
                PlanningToken~\citep{wang2024guiding}{,}
                RaLU~\citep{li2025reasoningaslogicunitsscalingtesttimereasoning}, leaf, text width=40em
            ]
        ]
        [
            \textbf{Hybrid Scaling}\\ (\S \ref{subsec:hybridsclaing})
            [
                MoA~\citep{wang2025mixtureofagents}{,}
                Tree of Thoughts~\citep{yao2023tree, zhang2024chain}{,}
                Graph of Thoughts~\citep{Besta2024graph}{,} Tree-Search\\
                \citep{chen2024tree}{,} 
                SoS~\citep{gandhi2024streams}{,}
                REBASE~\citep{wu2024scaling}{,} 
                OAIF~\citep{guo2024direct}{,}
                Beam-Search~\citep{guo2024direct}{,}M-\\
                CTS\citep{tian2024toward, zhang2024o1coder, gao2024interpretable, wan2024alphazero, chenalphamath}{,} 
                Journey Learning\citep{GAIR-o1p1}{,}A-\\
                daptiveAlloc\citep{snell2024scaling, ong2025routellm}{,}
                METAL\citep{li2025metalmultiagentframeworkchart}{,}
                rStar-Math\citep{guan2025rstarmath}{,}AtomThink\citep{xiang2024atomthinkslowthinkingframework}, leaf, text width=40em
            ]
        ]
        [
            \textbf{Internal Scaling}\\ (\S \ref{subsec:internalsclaing})
            [
                DeepSeek-R1~\citep{deepseek-r1}{,}
                OpenAI-o1\&o3~\citep{openai-o1,openai-o3}{,}
                Gemini Flash Thinking~\citep{geminithinking}{,}
                QwQ~\citep{qwq-32b-preview}{,}\\
                K1.5~\citep{kimi-k1.5}{,}
                3SUM~\citep{pfau2024lets}{,}
                OAIF~\citep{guo2024direct}{,}
                LIMO~\citep{ye2025limoreasoning}{,}
                 T1~\citep{hou2025advancing}{,}
                Distilled-o1\\~\citep{GAIR-o1p2}{,}
                RedStar~\citep{xu2025redstardoesscalinglongcot}{,}
                SKY-T1~\citep{skyt12025}{,}
                s1~\citep{muennighoff2025s1}{,} 
                ITT~\citep{hao2024training}, leaf, text width=40em
            ]
        ]
    ]
    [
        \textbf{How to Scale} \\ (\S \ref{sec:how2scale})
        [
            \textbf{Tuning} (\S \ref{subsec:tuning})
            [
                \textbf{Supervised} \\ \textbf{Finetuning} (\S \ref{subsubsec:sft})
                [
                    Distillation~\citep{muennighoff2025s1, GAIR-o1p2, xu2025redstardoesscalinglongcot, skyt12025, bespoke_stratos} \\
                    \citep{munkhbat2025selftrainingelicitsconcisereasoning, ye2025limoreasoning}{,}  
                    Synthesized Long CoT~\citep{hou2025advancing, yeo2025demystifying}{,} \\
                    Learning Reasoning Structure~\citep{li2025llmseasilylearnreason}{,} 
                    Long CoT warmup~\citep{kimi-k1.5}{ ,}
                    CFT~\citep{wang2025critiquefinetuninglearningcritique}, leaf2, text width=32em 
                ]
            ]
            [
                \textbf{Reinforcement}\\ \textbf{Learning} (\S \ref{subsubsec:rl})
                [
                    \textbf{Reward model-free}
                    [
                        Rule-Based~\citep{deepseek-r1}{,}
                        cDPO~\citep{lin2024critical}{,}
                        Focused-DPO\\
                        \citep{zhang2025focused}{,}
                        Selective DPO~\citep{gao2025principled}{,}
                        CPL~\citep{wang2024cpl}{,} \\
                        OREO~\citep{wang2024offline}{,} 
                        DAPO~\citep{liu2024improvingmultistepreasoningabilities}{,}
                        RFTT~\citep{zhang2025reasoning}{,} \\
                        SimPO~\citep{meng2024simpo}{,} 
                        DQO~\citep{ji2024enhancing}{,}
                        DAPO~\citep{yu2025dapo}{,} \\
                        VC-PPO~\citep{yuan2025s}{,} 
                        Light-R1~\citep{wen2025lightxi}{,} \textit{etc.}, leaf2, text width=24em
                    ]
                ]
                [   
                    \textbf{Reward model-based}
                    [
                        PPO~\citep{schulman2017proximalpolicyoptimizationalgorithms}{,}
                        RLOO~\citep{ahmadian2024back}{,} \\
                        GRPO~\citep{shao2024deepseekmath}{,}
                        DVPO~\citep{huang2025lean}{,} 
                        PRIME~\citep{cui2025process}{,}\\ 
                        REINFORCE++~\citep{hu2025reinforce++}{,}
                        SPPD~\citep{yi2025sppd}{,} \textit{etc.}, leaf2, text width=24em
                    ]
                ]
            ]
        ]
        [
            \textbf{Inference} (\S \ref{subsec:inference})
            [
                \textbf{Stimulation} (\S \ref{subsec:stimulation})
                [
                    \textbf{Prompt Strategy} 
                    [
                        Hint-infer~\citep{li2025startselftaughtreasonertools}{,}                        
                        Dipper~\citep{lau2024dipperdiversitypromptsproducing}{,}
                        EVA~\citep{ye2024evolvingalignmentasymmetricselfplay}{,} 
                        \\EvalPlan\citep{saha2025learningplanreason}{,} 
                        ReasonFlux~\citep{yang2025reasonflux}{,}~\citet{10460413}{,}\textit{ etc.}
                        , leaf2, text width=24em
                    ]
                ]
                [
                    \textbf{Decode Strategy} 
                    [
                        Filler Tokens~\citep{pfau2024lets}{,}
                        Budget Forcing~\citep{muennighoff2025s1}{,} \\
                        AFT~\citep{li2025draftsanswersunlockingllm}{,} 
                        Predictive-Decoding~\citep{ma2025nonmyopic}{,}\textit{ etc.}, leaf2, text width=24em
                    ]
                ]
                [
                    \textbf{Latent Strategy}
                    [
                        Coconut~\citep{hao2024training}{,}
                        CoDI~\citep{shen2025codicompressingchainofthoughtcontinuous}{,}
                        Heima~\citep{shen2025efficientreasoninghiddenthinking}{,} \\
                        Looped Transformers~\citep{saunshi2025reasoninglatentthoughtspower}{,}
                        LTV~\citep{kong2025scalablelanguagemodelsposterior}{,}\textit{ etc.}
                        , leaf2, text width=24em
                    ]
                ]
                [
                    \textbf{Self-Repetition}
                    [
                        Self-Consistency~\citep{wang2023selfconsistency}{,}
                        Self-Refine~\citep{madaan2023selfrefine}{,} DeCRIM\\
                        \citep{ferraz2024llmselfcorrectiondecrimdecompose}{,} CCE~\citep{zhang2025crowd}{,}
                        TreeBoN~\citep{qiu2024treebonenhancinginferencetimealignment}, leaf2, text width=24em
                    ]
                ]
                [
                    \textbf{Mixture-of-Model}
                    [   
                        MoA~\citep{wang2025mixtureofagents}{,}
                        RR-MP~\citep{he2025enhancingllmreasoningmultipath}{,}
                        BRAIN~\citep{chen2024braininspiredtwostageapproachenhancing}, leaf2, text width=24em
                    ]
                ]
            ]
            [
                \textbf{Verification} (\S \ref{subsec:verification})
                [
                    \textbf{Outcome}
                    [
                        Output Verification~\citep{cobbe2021training}{,} Generative Verifier~\citep{zhang2025generativeverifiersrewardmodeling}{,}\\Self-Reflection Feedback~\citep{li2025learningreasonfeedbacktesttime}{,} Discriminator~\citep{chen2024tree}{,}\\
                        OVM~\citep{yu2024ovm}{,} Heuristic~\citep{deepseek-r1}{,} Bandit~\citep{sui2025metareasonerdynamicguidanceoptimized}{,}\\
                        Functional~\citep{lee2025evolvingdeeperllmthinking}{,} 
                        XoT~\citep{liu2023plan}{,} 
                        WoT~\citep{zhang2024wrongofthoughtintegratedreasoningframework}, leaf2, text width=24em
                    ]
                ]
                [
                    \textbf{Process}
                    [
                        State Evaluator~\citep{yao2023tree,zhang2024chain}{,}
                        SIaM~\citep{yu2024siamselfimprovingcodeassistedmathematical}{,}\\
                        Deductive Verification~\citep{ling2023deductive}{,}
                        Self-Evaluator~\citep{xie2023selfevaluation}{,} \\
                        V-STaR~\citep{hosseini2024vstartrainingverifiersselftaught}{,}
                        Tool~\citep{li2025startselftaughtreasonertools}{,}
                        PoT~\citep{chen2023program}, leaf2, text width=24em
                    ]
                ]
            ]
            [
                \textbf{Search} (\S \ref{subsec:search})
                [
                    TreeSearch~\citep{yao2023tree,chen2024tree}{,}GraphSearch~\citep{Besta2024graph}{,}C-MSTS~\citep{lin2025leveragingconstrainedmontecarlo}{,}\\
                    MCTS~\citep{tian2024toward,zhang2024o1coder,gao2024interpretable, wan2024alphazero,chenalphamath}{,} SPaR\\
                    \citep{cheng2025sparselfplaytreesearchrefinement}{,}
                    REBASE~\citep{wu2024scaling}{,}
                    SoS~\citep{gandhi2024streams}{,} CoAT~\citep{pan2025coatchainofassociatedthoughtsframeworkenhancing}{,}Beam-\\
                    Search~\citep{guo2024direct,xie2023selfevaluation}{,}
                    Lookahead-Search~\citep{snell2024scaling, zhang2023planning}{,}\textit{ etc.}, leaf2, text width=32em
                ]
            ]
            [
                \textbf{Aggregation} (\S \ref{subsec:aggregation})
                [
                    \textbf{Selection}
                    [
                        Majority Voting\citep{wang2023selfconsistency, chen2024are}{,} BOND\citep{sessa2024bondaligningllmsbestofn}{,}\\
                        Filter Vote\citep{chen2024are}{,} 
                        Length-filtered Vote\citep{wu2025lessunderstandingchainofthoughtlength}{,} Best-of-N\\
                        \citep{irvine2023rewarding, song2024good}{,} Rejection Sampling~\citep{kimi-k1.5}{,} \textit{etc.} 
                        , leaf2, text width=24em
                    ]
                ]
                [
                    \textbf{Fusion}
                    [
                        BoN (weighted)~\citep{brown2024large}{,} Synthesize~\citep{wang2025mixtureofagents}{,} \textit{etc.}, leaf2, text width=24em
                    ]
                ]
            ]
        ]
    ]
    [
        \textbf{Where to Scale} \\ (\S \ref{sec:where2scale})
        [
            \textbf{Reasoning}\\\textbf{Intensive} (\S \ref{sec:reasoning})
            [
                \textbf{Math}
                [
                    AIME~\citep{aime25, guan2025rstarmath}{,}
                    CNMO~\citep{cnmo}{,}
                    NuminaMATH~\citep{numina_math_datasets}{,} OmniMath\\
                    \citep{gao2025omnimath}{,}
                    MATH~\citep{cobbe2021training, hendrycks2021measuring,guan2025rstarmath}{,} s1-prob-teasers\\
                    \citep{muennighoff2025s1}{,} 
                    GSM8K~\citep{guan2025rstarmath,zhang2024rest}{,} MATH500\citep{zhang2024rest}{,}\\
                    AMC~\citep{guan2025rstarmath}{,}
                    College Math~\citep{guan2025rstarmath}{,}
                    FrontierMath~\citep{glazer2024frontiermath}{,}\textit{ etc.}, leaf3, text width=32em
                ]
            ]
            [
                \textbf{Code}
                [
                    USACO~\citep{shi2024can}{,}
                    LiveCodeBench~\citep{jain2025livecodebench}{,}
                    CodeContests~\citep{Li2022competition}{,} Aider-Polyglot\\
                    \citep{aider}{,}SWE-bench\citep{jimenez2024swebench}{,}Codeforces\citep{codeforce}{,}CodeMind~\citep{liu2024codemindframeworkchallengelarge}{,}\textit{ etc.}
                    , leaf3, text width=32em
                ]
            ]
            [
                \textbf{Science}
                [
                    OlympicArena~\citep{huang2024olympicarena}{,}
                    OlympiadBench~\citep{he2024olympiadbench, guan2025rstarmath}{,} TheoremQA\\
                    \citep{chen2023theoremqa}{,}
                    JEEBench~\citep{arora2023have}{,}    
                    GPQA~\citep{rein2024gpqa}{,}
                    SciEval~\citep{sun2024scieval}{,}\\
                    Miverva~\citep{lewkowycz2022solving}{,}
                    SciBench~\citep{zhang2024rest}{,} HLE~\citep{phan2025humanity}{,}\textit{ etc.}\\
                    , leaf3, text width=32em
                ]
            ]
            [
                \textbf{Game \& Strategy}
                [
                    SysBench~\citep{aime25}{,} Points24~\citep{yao2023tree,zhai2024finetuninglargevisionlanguagemodels}{,} TravelPlan~\citep{xie2024travelplannerbenchmarkrealworldplanning}{,}\textit{ etc.}, leaf3, text width=32em
                ]
            ]
            [
                \textbf{Medical}
                [
                    SysBench{,} JMLE-2024~\citep{chen2025benchmarkinglargelanguagemodels}{,}
                    Medbullets~\citep{chen2025benchmarkinglargelanguagemodels}{,}
                    MedQA~\citep{jin2020diseasedoespatienthave}{,}\textit{ etc.}, leaf3, text width=32em
                ]
            ]
        ]
        [   
            \textbf{Others} (\S \ref{sec:generalpurpose})
            [
                \textbf{General}
                [
                    AGIEval~\citep{zhong2024agieval}{,} 
                    MMLU-Pro~\citep{wang2024mmlupro}{,}  
                    Gaokao~\citep{gaokao, guan2025rstarmath}{,}\\
                    Kaoyan~\citep{kaoyan}{,} CMMLU~\citep{li2024cmmlumeasuringmassivemultitask}{,} LongBench~\citep{bai2024longbenchbilingualmultitaskbenchmark}{,} ARC-AGI~\citep{chollet2019measureintelligence}{,}\textit{ etc.}, leaf3, text width=32em
                ]
            ]
            [
                \textbf{Agents}
                [
                    WebShop~\citep{yao2023webshop}{,} 
                    WebArena~\citep{zhou2023webarena}{,} 
                    SciWorld~\citep{wang2022sciworld}{,} 
                    WebVoyager\\
                    \citep{he2024webvoyagerbuildingendtoendweb}{,} TextCraft~\citep{prasad2024adaptasneededdecompositionplanning}{,} TAU-bench~\citep{yao2024taubenchbenchmarktoolagentuserinteraction}{,} BCFL~\citep{berkeley-function-calling-leaderboard}{,}\textit{ etc.}, leaf3, text width=32em
                ]
            ]
            [
                \textbf{Knowledge}
                [
                    SimpleQA~\citep{wei2024measuringshortformfactualitylarge}{,} 
                    C-SimpleQA~\citep{he2024chinesesimpleqachinesefactuality}){,}
                    FRAMES~\citep{krishna2025factfetchreasonunified}{,}\textit{ etc.}, leaf3, text width=32em
                ]
            ]
            [
                \textbf{Open-Ended}
                [
                    AlpacaEval2.0~\citep{dubois2024lengthcontrolledalpacaevalsimpleway}{,} 
                    ArenaHard~\citep{li2024crowdsourceddatahighqualitybenchmarks}{,} 
                    IF-Eval~\citep{zhou2023instructionfollowing}{,} Chatbot Arena\\
                    \citep{NEURIPS2023_91f18a12}{,}
                    C-Eval~\citep{huang2023ceval}{,} FollowBench~\citep{jiang2024followbenchmultilevelfinegrainedconstraints}{,}\textit{ etc.}, leaf3, text width=32em
                ]
            ]
            [
                \textbf{Evaluation}
                [
                    RewardBench~\citep{lambert2024rewardbenchevaluatingrewardmodels}{,}
                    JudgeBench~\citep{tan2025judgebenchbenchmarkevaluatingllmbased}{,}
                    RMBench~\citep{liu2024rmbenchbenchmarkingrewardmodels}{,} \\
                    PPE~\citep{frick2024evaluaterewardmodelsrlhf}{,}
                    RMB~\citep{zhou2025rmbcomprehensivelybenchmarkingreward}{,}\textit{ etc.}, leaf3, text width=32em
                ]
            ]
            [
                \textbf{Multi-Modal}
                [
                    MMMU~\citep{yue2024mmmu}{,} 
                    MATH-Vision~\citep{wang2024measuring}{,}
                    MathVista~\citep{lu2024mathvista}{,} LLAVA-Wild\\
                    \citep{liu2023visual}{,}
                    MM-Vet~\citep{yu2024mm}{,}
                    MMBench~\citep{liu2024mmbench}{,}
                    MMMU~\citep{yue2024mmmu}{,} \\
                    CVBench~\citep{tong2024cambrian}{,}
                    MMStar~\citep{chen2024we}{,}
                    CHAIR~\citep{rohrbach2018object}{,}\textit{ etc.}, leaf3, text width=32em
                ]
            ]
        ]
    ]
    [
        \textbf{How Well to}  \\ \textbf{Scale} (\S \ref{sec:howwell2scale})
        [
            \textbf{Accuracy} \\ (\S \ref{subsec:performance})
            [
               Pass@1~\citep{deepseek-r1, kimi-k1.5}{, }Pass@k\citep{chen2021evaluating, brown2024large}{, }WinRate\citep{deepseek-r1, hou2025advancing}\\Cons@k~\citep{deepseek-r1, zeng2025revisiting}{, }\textit{etc.}, leaf4, text width=40em
            ]
        ]
        [
            \textbf{Efficiency} \\ (\S \ref{subsec:efficiency})
            [
            Token Cost~\citep{welleck2024decoding, aytes2025sketchofthoughtefficientllmreasoning}{, }
            FLOPs-based Efficiency Analysis~\citep{kaplan2020scalinglawsneurallanguage, snell2024scaling}{, }\\
            KV Cache size~\citep{hooper2025etsefficienttreesearch}{, }
            Underthinking score~\citep{wang2025thoughtsplaceunderthinkingo1like}{,}
            \textit{ etc.}, leaf4, text width=40em
            ]
        ]
        [
            \textbf{Controllability}\\ (\S \ref{subsec:controllability})
            [
               Control Metric~\citep{muennighoff2025s1}{,} Length Deviation~\citep{aggarwal2025l1}{,}$k$-$\epsilon$ Controllability~\citep{bhargava2024whatsmagicwordcontrol}{,}\textit{ etc.}, leaf4, text width=40em
            ]
        ]
        [
            \textbf{Scalability}\\ (\S \ref{subsec:scalability})
            [
               Scaling Metric~\citep{muennighoff2025s1}{,} Scaling Curves (Accuracy vs. Compute)~\citep{aggarwal2025l1, teng2025atom}{,}\textit{ etc.}, leaf4, text width=40em
            ]
        ]
    ]
]
\end{forest}
}
\vspace{-1em}
\caption{Taxonomy of research in Test-time Scaling that consists of what, how, where, and how well to scale.}
\label{categorization_of_reasoning}
\end{figure}
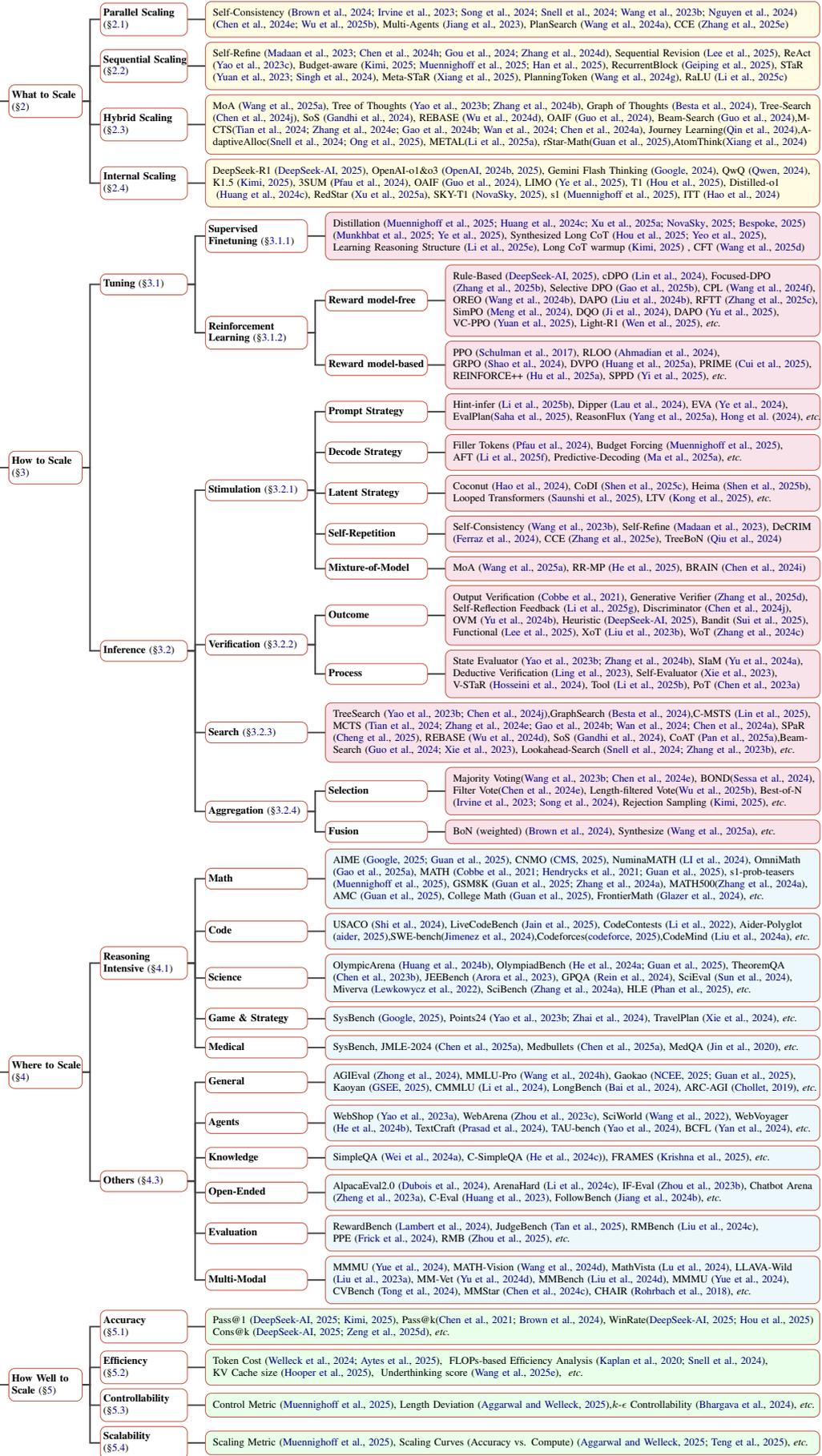

\section{What to Scale}
\label{sec:what2scale}
``What to scale'' refers to the specific form of \TTS that is expanded or adjusted to enhance an LLM's performance during inference. 
When applying \TTS, researchers typically choose a specific ``what to scale'' based on an empirical hypothesis, aiming to achieve performance gains. For example, some researchers hypothesize that longer CoTs improve complex reasoning, leading them to enforce longer outputs from LLMs. Others leverage the self-consistency principle, assuming that generating multiple solutions to a reasoning task increases the likelihood of reaching the correct answer.

\subsection{Parallel Scaling}
\label{subsec:parallelsclaing}
LLMs typically generate a single response per query. \textit{Parallel scaling} improves test-time performance by generating multiple outputs in parallel and then aggregating them into a final answer. 
Formally, consider a problem set \( \mathcal{P} \) and a collection of models \( m \in \{1, \dots, M\} \). Each model generates \( k_m \) candidate responses for a given problem \( p \in \mathcal{P} \), producing a set of sampled solutions $\mathcal{S}$:
\begin{align}
    \mathcal{S} &= \{ s_{m,i} \mid m \leq M,\, i \leq k_m \},
    \quad \Rightarrow \quad
    \left( \exists \hat{s} \right) \, \hat{s} = A(s_{1,1}, \dots, s_{M,k_M}) \text{ is correct}.
\end{align}

Here, \( A \) is the aggregation function that derives a final response from the set \( \mathcal{S} \). The effectiveness of parallel scaling depends on both \textbf{coverage}—the likelihood of generating at least one correct response—and \textbf{aggregation quality}, which determines whether a correct response is successfully identified.
This approach is supported by both theory and intuition: cognitive science research~\citep{Stanovich_West_2000} suggests that complex problems often allow multiple valid solution paths, and increasing the number of generated responses improves the chance of finding a correct one~\citep{li2025stesttimescaling}. Empirically, this relationship is often log-linear with respect to compute~\citep{brown2024large}.

We categorize parallel scaling into two common forms based on different sources of coverage: (1) repeated sampling from a single model and (2) sampling across multiple models. 
Furthermore, there are some additional techniques to enhance solution diversity and reliability, such as hyperparameter adjustments (\eg, sampling temperature~\citep{renze2024effectsamplingtemperatureproblem} to control output variability) and input modifications (\eg, prompt rephrasing~\citep{lambert2025tulu3pushingfrontiers} to elicit diverse responses). 



\subsection{Sequential Scaling}
\label{subsec:sequentialsclaing}
\textit{Sequential scaling} involves explicitly directing later computations based on intermediate steps. 
Unlike parallel methods, sequential scaling updates intermediate states iteratively. We denote the partial solution states (subproblem results, or initial drafts) by $n_1, n_2, \dots, n_T$, with each new state $n_{t+1} = R(n_t, p)$ incorporating both the previous state and the problem context.
Because many problems require deliberation rather than immediate pattern matching, single-pass `System 1'~\citep{yu2024distilling21}-style generation often fails on complex reasoning tasks. Iterative methods emulate a `System 2' approach, breaking down and refining the solution step by step.

Early work like chain-of-thought prompting~\citep{wei2022chain} motivated solve the problem step-by-step, $n_{t+1} = \text{AppendStep}(n_t, \text{new reasoning step})$, leading to approaches that refine responses~\citep{madaan2023selfrefine}, $n_{t+1} = \text{Refine}(n_t)$, or break down problems systematically~\citep{zhou2023leasttomostpromptingenablescomplex, zelikman2022star}, $n_{t+1} = \text{IntegrateSub}\bigl(n_t,\, \text{solution to next subproblem}\bigr)$. Subsequent studies show that iterative revision~\citep{chen2024teaching, gou2024critic,chen2025iterativedeepeningsamplinglarge, snell2024scaling} triggers self-correction, improving accuracy on challenging tasks
In practice, real-world tasks often demand more flexible and potentially non-linear reasoning paths, suggesting that purely sequential approaches, while effective, may be only one part of a broader solution.

\subsection{Hybrid Scaling}
\label{subsec:hybridsclaing}
\textit{Hybrid scaling} exploits the complementary benefits of parallel and sequential scaling. Parallel scaling mitigates the risk of the model missing the correct line of thought by casting a wide net, while sequential scaling allows deep exploration of a line of reasoning once it seems promising. 
Formally, let $\mathcal{F}_t$ be the set of candidate solutions at iteration $t$. 
Each iteration expands these candidates in parallel with an expansion function $\mathcal{E}$ and sequentially filters them with a selection function $\mathcal{S}$:
\begin{align}
  \mathcal{F}_{t+1} \;=\;
  \mathcal{S}\Bigl(\bigcup_{s \in \mathcal{F}_t} \mathcal{E}(s)\Bigr).
\end{align}

After $T$ iterations, an aggregator $A$ selects the final solution $\hat{s} \in \mathcal{F}_T$. 
From a cognitive standpoint, such a combination mirrors how human problem-solvers generate multiple hypotheses (divergent thinking) and then refine/evaluate them (convergent thinking).
Classic search algorithms (\eg, iterative deepening~\citep{chen2025iterativedeepeningsamplinglarge} and beam search~\citep{snell2024scaling}) embody this strategy by balancing exploration and exploitation.

Recent work expands on this idea.
Tree-of-Thoughts (ToT)~\citep{yao2023tree} branches at decision points, exploring multiple reasoning paths before pruning to a single sequence. Follow-up methods, such as Graph-of-Thoughts~\citep{Besta2024graph}, Algorithm-of-Thought~\citep{sel2024algorithm}, Forest-of-Thought~\citep{bi2024forest}, Monte Carlo Tree Search (MCTS)~\citep{lin2025leveragingconstrainedmontecarlo}, and multi-agent reasoning~\cite{wang2025mixtureofagents,chen2024routerdcquerybasedrouterdual}, leverage similar but more complex hybrid patterns. For instance, multiple LLMs can debate or verify each other’s answers \citep{liang2024encouraging, schaul2024boundlesssocraticlearninglanguage}, while ``journey learning'' and ``tool-augmented reasoning''~\citep{li2025startselftaughtreasonertools} emphasize capturing full reasoning trajectories.

\subsection{Internal Scaling}
\label{subsec:internalsclaing}
\textit{Internal scaling} elicits a model to autonomously determine how much computation to allocate for reasoning during testing within the model's internal parameters instead of depending on external human-guided strategies.
Formally, we update an initial model $M_{0}$ to a new model $M_{1}$ via a training procedure, $\Phi : (M_{0}, \mathcal{D}) \;\mapsto\; M_{1}$, on data $\mathcal{D}$ that includes multi-step reasoning tasks (\eg, long CoT examples produced by external scaling~\citep{GAIR-o1p1}).
Surprisingly, employing outcome-oriented reward modeling~\citep{deepseek-r1, openai-o1} for RL enables the model to extend its reasoning process autonomously. 

At test time, $M_{1}$ generates a sequence of internal states $z_{1}, z_{2}, \dots, z_{T}$ via
\begin{align}
    z_{t+1} = f_{\theta}(z_t), 
  \quad
  \mathrm{stop}(z_t) = \pi_{\theta}(z_t).
\end{align}
The model’s learned policy $\pi_{\theta}$ controls when to halt.
This internal feedback loop can lead to emergent behaviors—such as more detailed reasoning chains or self-evaluation steps—without any external prompts or multi-call orchestration. In practice, internal scaling often rivals or surpasses standard techniques, thanks to its ability to focus computational effort on a single, coherent reasoning trajectory.

\section{How to Scale}
\label{sec:how2scale}

\subsection{Tuning-based Approaches}
\label{subsec:tuning}

To activate a model’s ability to devote cost at test time, directly tuning its parameters is an effective strategy. This includes two approaches:
1) Supervised Finetuning (SFT): Training an LLM via next-token prediction on synthetic or distilled long CoTs enables it to imitate and internalize structured reasoning patterns, effectively learning to think through complex problems. By mimicking extended rationales, SFT reduces the reliance on explicit prompting at inference time. 
2) Reinforcement Learning (RL): By leveraging feedback from a reward model on inference tasks, the policy model is automatically updated. Although no supervised data is introduced, the model autonomously generates long CoT reasoning while ensuring reliable answers.
We divide the RL for internal scaling works into two perspectives. The reward model-based methods and the reward model-free methods. 

\subsubsection{Supervised Finetuning (SFT)}
\label{subsubsec:sft}
Training an LLM via next-token prediction on synthetic or distilled long CoTs enables it to internalize structured reasoning patterns and effectively ``think'' through complex problems. By mimicking extended rationales, SFT reduces the reliance on explicit prompting at inference time. This will include three subsections: (1) Imitation, describing techniques like MCTS used to generate CoT-style demonstrations for fine-tuning, (2) Distillation, summarizing how student models are trained using outputs from stronger models (\eg, o1, R1), and (3) Warmup, stabilizing learning and aligning the model’s behavior to produce useful step-by-step reasoning.

\paragraph{Imitation}
A prominent approach to enhancing LLM reasoning via SFT is to generate long CoT demonstrations using test-time ``planner'' algorithms and then fine-tune the model to imitate those demonstrations. For example, STaR~\citep{zelikman2022star} uses the model itself to generate step-by-step solutions for a given problem and filters for correct outcomes, treating the verified solutions as new demonstrations to fine-tune. More structured search has been applied to generate even higher-quality traces: ReST-MCTS~\citep{zhang2024rest} integrates an MCTS planner (guided by a learned value model) to explore the space of possible reasoning steps; the model is subsequently fine-tuned on these search-generated traces, \ie, it learns to imitate the successful reasoning trajectories discovered by the planner.

\paragraph{Distillation}
While the imitation approach uses a model’s own intermediate outputs for improvement, distillation techniques aim to transfer the capabilities of a stronger model (or ensemble of models) into a target model via supervised learning. As reported by~\citet{muennighoff2025s1,li2025llmseasilylearnreason}, a 32B model trained on a curated sample set generated by a top-tier reasoner was able to solve competition-level math problems nearly as well as the teacher, indicating successful distillation of reasoning.

\paragraph{Warmup}
SFT warmup~\citep{luong2024reftreasoningreinforcedfinetuning} refers to an initial SFT phase applied to an LLM after its unsupervised pretraining but before other post-training steps like RL. This stage stabilizes subsequent training by providing a well-initialized model that adapts better to preference optimization and avoids instability due to ungrounded behavior~\citep{zeng2025itoolboostingtooluse}. Effective SFT warmup is characterized by several key elements: (i) the use of high-quality, task-relevant datasets~\citep{luong2024reftreasoningreinforcedfinetuning}; (ii) short duration; (iii) a tailored learning rate schedule~\citep{pareja2024unveilingsecretrecipeguide}. Technically, SFT warmup is often integrated with methods like rejection sampling~\citep{pareja2024unveilingsecretrecipeguide}—which uses warm-started models to generate high-quality data for further training.

\subsubsection{Reinforcement Learning (RL)}
\label{subsubsec:rl}

\paragraph{Reward model-free.} Recent advancements in RL and preference optimization have significantly enhanced the performance of large language models, particularly in reasoning and problem-solving tasks. A key innovation in this domain is the introduction of RL with verifiable reward by DeepSeek R1~\citep{deepseek-r1}, which leverages rule-based reward mechanisms to optimize models efficiently and reliably. This approach has sparked growing interest among researchers working on large models, as it addresses challenges such as sparse rewards and training instability by providing dense feedback for policy optimization.
Several methods have been developed to improve exploration and accuracy in reasoning tasks through preference optimization. For instance, cDPO~\citep{lin2024critical}, CPL~\citep{wang2024cpl}, Focused-DPO~\citep{zhang2025focused}, DAPO~\citep{liu2024improvingmultistepreasoningabilities}, and RFTT~\citep{zhang2025reasoning} prioritize critical or error-prone areas, enhancing internal scaling and reasoning accuracy. Additionally, Selective DPO~\citep{gao2025principled} emphasizes the importance of aligning data difficulty with model capacity by filtering out overly challenging examples, further refining the training process.
VC-PPO~\citep{yuan2025s} investigates the failure of PPO for the long CoT task and uses a pre-trained value model to achieve better results. 
Light-R1~\citep{wen2025lightxi} proposes a curriculum training framework for increasing data difficulty combined with multi-staged post-training. 
SimPO~\citep{meng2024simpo} uses the average log probability of a sequence as the implicit reward and removes the reference model in DPO.

In the realm of mathematical problem-solving, DQO~\citep{ji2024enhancing} and OREO~\citep{wang2024offline} propose novel value function optimization techniques, demonstrating improvements in model performance. 
DAPO~\citep{yu2025dapo} leverages dynamic sampling for large-scale RL systems. 
These advancements are complemented by a range of open-source training frameworks that have equipped researchers and developers with tools to optimize training and enhance inference. Early frameworks like SimpleRL~\citep{zeng2025simplerl} and DeepScaler~\citep{deepscaler2025} quickly replicated the technology stack of DeepSeek R1. Furthermore, SimpleRL-Zoo~\citep{zeng2025simplerlzoo} presents more experimental details about SimpleRL.  Others, such as X-R1~\citep{xr12025} and TinyZero~\citep{tinyzero}, focus on delivering an intuitive and cost-effective user experience. Notably, Open-Reasoner-Zero~\citep{OpenReasonerZero2025} replicated the DeepSeek R1-zero training scheme using a 32B model, achieving comparable performance.
Further advancements in RL for internal scaling have been facilitated by frameworks like OpenR~\citep{wang2024openr}, OpenRLHF~\citep{hu2024openrlhf}, OpenR1~\citep{openr1}, Logic-RL~\citep{xie2025logic} and AReaL\citep{areal2025}. These frameworks have enhanced the replication of internal scaling and, through open-source sharing, accelerated academic research progress. 
The above developments not only address key challenges in RL but also pave the way for more efficient and reliable model training and deployment.

\paragraph{Reward model-based.}
With a Bradley-Terry model~\citep{zheng2023secrets} optimized by human preference as the reward model, PPO~\citep{schulman2017proximalpolicyoptimizationalgorithms} stands as one of the most influential algorithms with its efficiency and stability and is widely used for internal scaling. 
Building upon PPO, ReMax~\citep{li2023remax} introduces variance reduction techniques along with REINFORCE~\citep{sutton1999policy} and RLOO~\citep{ahmadian2024back} methods. This eliminates the need for additional value models in PPO, reduces over four hyperparameters, lowers GPU memory usage, and speeds up the training process.
GRPO~\citep{shao2024deepseekmath} replaces traditional value models with improved sampling strategies. This significantly accelerates the learning process and achieves performance comparable to GPT-4 in mathematics.  REINFORCE++~\citep{hu2025reinforce++} further simplifies GRPO and enhances its training.
DVPO~\citep{huang2025lean} presents a streamlined framework, substituting the reward model with a pre-trained global value model and removing the dependency between the actor and critic. PRIME~\citep{cui2025process} integrates the SFT model as a PRM within a unified RL framework, allowing online updates through policy rollouts and outcome labels via implicit process rewards. SPPD~\citep{yi2025sppd} utilizes process preference learning with a dynamic value margin for self-training.
Recently, several works have focused on other challenges of existing reward model-based methods. UGDA~\citep{sun2025uncertain} leverages the uncertainty and influence of samples during PPO training and iteratively refines the reward model. VinePPO~\citep{kazemnejad2024vineppo} exploits the flexibility of language environments to compute unbiased Monte Carlo-based estimates, avoiding the need for large value networks. LCPO~\citep{aggarwal2025l1} focuses on optimizing accuracy and adherence to user-specified length constraints for reasoning tasks. Rest-MCTS*~\citep{zhang2024rest} uses tree-search-based RL to bypass per-step manual annotation typically required for training process rewards.
These advancements and refinements in algorithms continue to drive the field of reinforcement learning for internal scaling, offering more effective tools and methods for solving complex problems.



\subsection{Inference-based Approaches}
\label{subsec:inference}

\begin{figure}[!htbp]
    \centering
    \includegraphics[width=0.98\linewidth]{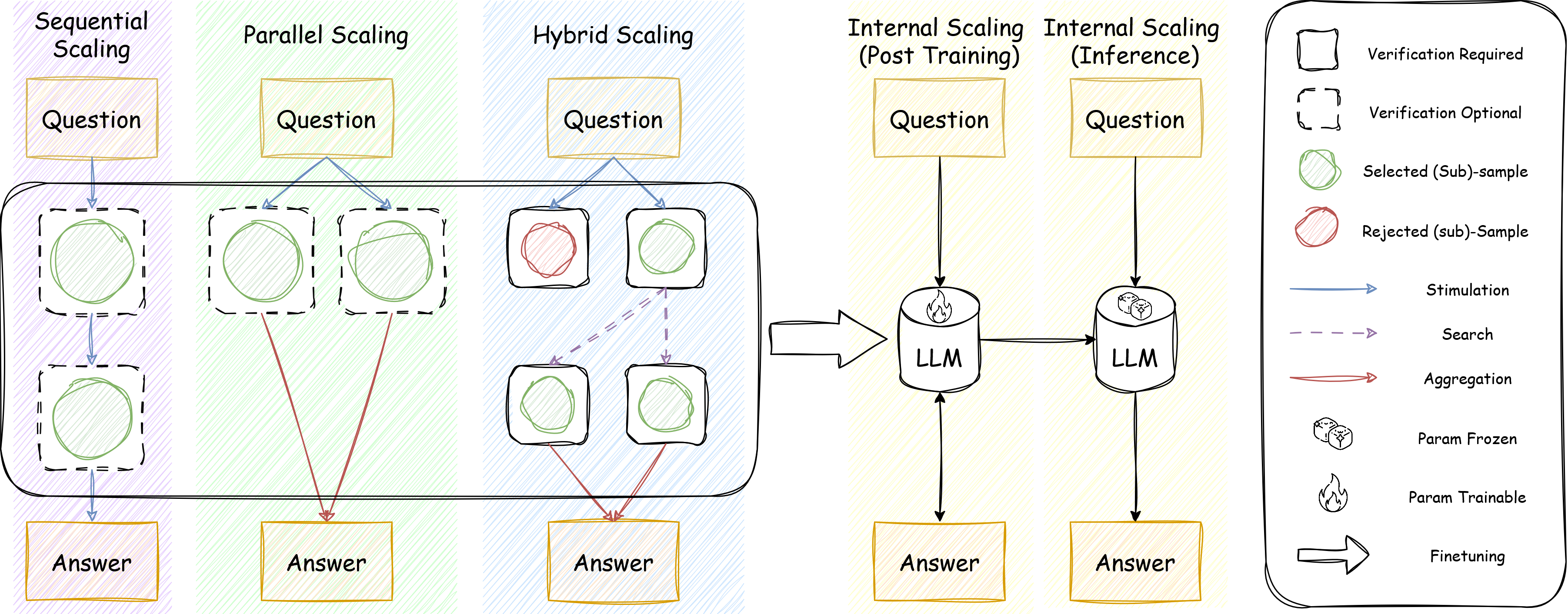}
    \caption{A Visual Map and Comparison: From \textit{What to Scale} to \textit{How to Scale}.}
    \label{fig:illustration}
\end{figure}

Unlike training-based approaches, which adjust the model’s parameters offline, inference-based approaches dynamically adjust computation during deployment. This paradigm includes four essential components: (i) \textit{Stimulation}, which encourages the model to generate longer or multiple candidate outputs; (ii) \textit{Verification}, which filters or scores outputs based on correctness or other criteria; (iii) \textit{Search}, which systematically explores the sample space; and (iv) \textit{Aggregation}, which consolidates multiple outputs into the final output. These four components are often used in combination to allocate test-time computation more effectively and boost performance on complex reasoning tasks. In the following sections, we provide detailed discussions of each component.

\subsubsection{Stimulation}
\label{subsec:stimulation}

Stimulation techniques are the first step in encouraging the model to allocate more computation to thinking. It basically stimulates the LLM to generate (i) longer samplers and (ii) more samples instead of generating single and short samples via naive prompting. This includes several key approaches:

\paragraph{Prompt Strategy.} Instead of allowing the model to generate an answer directly, one way to stimulate the scaling of LLM during test time is through the prompt. This behavior requires the backbone LLM's ability to follow instructions. For instance, prompts can guide the model toward step-by-step reasoning. Simple modifications such as adding explicit instructions (\eg, ``Please think step by step.'') can improve the model’s ability to break down complex problems into intermediate steps~\citep{lightman2023let}. This strategy ensures more deliberate and structured thought generation by shaping the reasoning process at the input level. Other techniques such as ~\cite{wei2022chain, ranaldi2025improvingchainofthoughtreasoningquasisymbolic} also rely on explicitly stating the requirements in the prompt to stimulate samples during the \TTS.

\paragraph{Decode Strategy} Rather than passively accepting the model’s default output behavior, this approach modifies the decoding process to encourage LLM to generate longer, more detailed samples adaptively. Techniques such as 
injecting filler token~\citep{pfau2024lets}, adaptively injecting predefined injection phrase~\citep{jin2020diseasedoespatienthave}, forcing scaling budget~\citep{muennighoff2025s1}, enforcing intermediate generation~\citep{li2025draftsanswersunlockingllm}, or predictive decoding~\citep{ma2025nonmyopic} allow the model to modify its distribution progressively. Enforcing extended reasoning at the output level enables the model to think longer and generate more comprehensive solutions without requiring additional external guidance.

\paragraph{Latent Strategy}  Unlike strategies that rely on token-level instructions or output expansion, latent strategies encourage deeper or recurrent thinking within the hidden representations themselves, effectively scaling up test-time computation through continuous internal states. For example, \citet{hao2024training} propose a paradigm where the model completes reasoning steps entirely in hidden space before producing the final answer; \citet{kong2025scalablelanguagemodelsposterior} introduce a latent-thought framework that conditions text generation on an inferred latent variable to guide more thorough or expansive reasoning, while \citet{shen2025codicompressingchainofthoughtcontinuous} show that compressing CoT into continuous embeddings can preserve intermediate reasoning fidelity without lengthy textual traces. Other approaches~\citep{saunshi2025reasoninglatentthoughtspower} harness \emph{looped} or \emph{recurrent} inference to repeatedly refine hidden states, effectively unfolding multiple ``thinking iterations'' in a single forward pass.

\paragraph{Self-Repetition Strategy} Apart from generating longer samples, another way to stimulate the LLM is to generate multiple samples instead of individual ones. One commonly adopted strategy is to prompt the LLM repeatedly during the decoding stage, commonly known as self-repetition~\citep{wang2023selfconsistency}. Another strategy is to prompt the LLM sequentially, in order to mimic refinement process~\citep{madaan2023selfrefine} or correlation under constraint~\citep{ferraz2024llmselfcorrectiondecrimdecompose}. 

\paragraph{Mixture-of-Model Strategy} Gathering the ``wisdom of the crowd'' can move beyond repeated sampling from a single model to coordinated sampling across multiple models. These LLMs can play either homogeneous roles~\citep{wang2025mixtureofagents} or heterogeneous roles~\citep{chen2024braininspiredtwostageapproachenhancing,he2025enhancingllmreasoningmultipath} during the process. By harnessing diverse perspectives, such multi-model strategy not only increases the coverage of possible solutions but also improves the system’s overall robustness. 

\begin{table*}[!htbp]
    \centering
    \resizebox{0.98\textwidth}{!}{
    \begin{tabular}{lll}
    \toprule
        \rowcolor{gray!10}
        \textbf{Category} & \textbf{Approach} & \textbf{Approach Description} \\
    \midrule
        & CoT~\citep{wei2022chain} & Contains a series of intermediate reasoning steps in prompts \\
        & Step-by-step~\citep{lightman2023let} & Stimulate step-by-step thinking via prompt \\
        & QuaSAR~\citep{ranaldi2025improvingchainofthoughtreasoningquasisymbolic} & Decompose CoT into Quasi-Symbolic Language \\
        & CoD~\citep{xu2025chaindraftthinkingfaster} & Generate concrete representations and distill into concise equation \\
        & Hint-infer~\citep{li2025startselftaughtreasonertools} & Inserting artificially designed hints in the prompt \\
        & Think~\citep{li2025startselftaughtreasonertools} & Prompt LLM with ``Think before response`` \\
        \multirow{-7}{*}{\textbf{Prompt}}
        & Think About World~\citep{jin2024impact} & Prompt LLM with ``Think About the World`` to enforce larger inference \\
    \midrule
        \rowcolor{gray!10}
        & Filler-token~\citep{pfau2024lets} & uses arbitrary, irrelevant filler tokens before answering \\
        \rowcolor{gray!10}
        & Budget-forcing~\citep{muennighoff2025s1} & suppress the generation of the end-of-thinking token \\
        \rowcolor{gray!10}
        & AFT~\citep{li2025draftsanswersunlockingllm} & iteratively aggregating proposals and aggregate for future proposals \\
        \rowcolor{gray!10}
        & Predictive-Decoding~\citep{ma2025nonmyopic} & re-weight decoding distribution given evaluation of foresight\\
        \rowcolor{gray!10}\multirow{-5}{*}{\textbf{Decode}} 
        & Adaptive Injection~\citep{jin2025wellthinkingenhancingllm} & Injecting a predefined injection phrase under certain condition \\
    \midrule
        & Coconut~\citep{hao2024training} & Perform chain-of-thought in hidden space without explicit token generation \\
        & CoDI~\citep{shen2025codicompressingchainofthoughtcontinuous} & Compress chain-of-thought into continuous vectors via self-distillation \\
        & Looped (Recurrent) Transformers~\citep{saunshi2025reasoninglatentthoughtspower} & Unroll model depth at inference by repeatedly refining hidden states \\
        & Heima~\citep{shen2025efficientreasoninghiddenthinking} & Encode each reasoning step into a single latent token to reduce output length \\
        \multirow{-5}{*}{\textbf{Latent}}
        & LTV~\citep{kong2025scalablelanguagemodelsposterior} & Introduce a latent thought variable to guide text generation \\
    \midrule
        \rowcolor{gray!10}
        & Self-Repetition~\citep{wang2023selfconsistency} & prompt LM in parallel \\
        \rowcolor{gray!10}
        & Self-Refine~\citep{madaan2023selfrefine} & Naively prompt LM to iteratively refine answer \\
        \rowcolor{gray!10}\multirow{-3}{*}{\textbf{Self-Repetition}}
        & DeCRIM~\citep{ferraz2024llmselfcorrectiondecrimdecompose} & Self-correlation for multi-constrained instruction following \\
    \midrule
        & MoA~\citep{wang2025mixtureofagents} & Prompt different models in parallel and iteratively improve \\
        & RR-MP~\citep{he2025enhancingllmreasoningmultipath} & Propose Reactive and Reflection agents to collaborate \\
        & BRAIN~\citep{chen2024braininspiredtwostageapproachenhancing} & Propose frontal \& parietal lobe model to inspire brain \\
        \multirow{-4}{*}{\textbf{Mixture-of-Model}}
        & Collab~\citep{chakraborty2025collab} & Propose decoding strategies to leverage multiple off-the-shelf aligned LLM policies \\
    \bottomrule
    \end{tabular}}
    \caption{Summary of Certain Stimulation Techniques.}
    \label{tab:stimulation}
\end{table*}

\subsubsection{Verification}
\label{subsec:verification}
Verifying the correctness and consistency of LLM during the test-time scaling is also crucial. The verification process plays an important role in the test-time scaling, as a solid verification process can be adapted to: 
\begin{itemize}
    \item directly selects the output sample among various ones, under the \textit{Parallel Scaling} paradigm;
    \item guides the stimulation process and determines when to stop, under the \textit{Sequential Scaling} paradigm;
    \item serves as the criteria in the search process, which we will discuss in Section~\ref{subsec:search};
    \item determines what sample to aggregate and how to aggregate them, e.g., weights, which we will discuss in Section~\ref{subsec:aggregation}.
\end{itemize}
Usually, there are two types of verifications, as shown below:

\paragraph{Outcome Verification.}
Outcome verification plays a crucial role in ensuring the correctness and consistency of generated outputs. Common approaches include using a separate verifier model to score multiple candidate answers (\eg,\citet{cobbe2021training}), employing self-consistency, voting mechanisms~\citep{wang2023selfconsistency} and discriminator LM~\citep{chen2024tree} and leveraging tool-assisted~\citep{gou2024critic} or heuristic checks~\citep{deepseek-r1} in domains such as math and code generation. 
For specific task problems, such as trip planning, functional scoring~\citep{lee2025evolvingdeeperllmthinking} is also adopted for verifying the proposed plans. 
Instead of formulating the outcome verification as a classification problem, \citet{zhang2025generativeverifiersrewardmodeling} exploits the generative ability of LLM and proposes to reformulate the outcome verification process as a next-token prediction task.
\citet{li2025learningreasonfeedbacktesttime} formulate the feedback utilization as an optimization problem and adaptive propagate information between samples.

Apart from single criteria, certain outcome verification approaches verify the quality of the simulated samples from multiple perspectives. 
\citet{liu2023plan} conducts both (i) passive verification from external tools and (ii) active verification via a rethinking mechanism to justify each sample.
\citet{zhang2024wrongofthoughtintegratedreasoningframework} follows a similar idea and proposes to verify each sample from three aspects: Assertion, Process, and Result.
\citet{lifshitz2025multiagent} further extends the number of verification agents to an arbitrary number and decouples the semantic criteria with verification agents. 
\citet{parmar2025plangenmultiagentframeworkgenerating} and \citet{saadfalcon2024archonarchitecturesearchframework} also propose a verification agent to score each sample considering various factors, respectively. \citet{saadfalcon2024archonarchitecturesearchframework} additionally proposes a unit test-based verification approach.
We provide a detailed technical categorization in the Appendix~\ref{app:outcome_verification}.


\paragraph{Process Verification.}
Process verification approaches verify the sample outcomes and the process of obtaining such an outcome. They are commonly adopted in tasks with formal, deductive processes, such as reasoning, coding, or mathematics. They are also known as the process reward model (PRM) or state verification. \citet{lightman2023let} processes to train a PRM as a step-level verification on mathematical tasks. \citet{yao2023tree} processes an LM-based state verifier as guidelines for searching the samples under the tree structure. \citet{zhang2024chain} further tunes these preference data into LLM and enables CoT structure during test time. Instead of training an external verifier, \citet{xie2023selfevaluation} prompts the same LM to evaluate the current step given all previous ones. \citet{hosseini2024vstartrainingverifiersselftaught} proposes to train the verifier with both accurate and inaccurate generated data.
Although LM-based process verifiers can be easily integrated, they may yield unreliable verification, especially for complex problems with long processes. \citet{ling2023deductive} decomposes the verification process in a deductive manner. Hence, the verifier only needs to verify a few statements within the long thought chain. \citet{yu2024siamselfimprovingcodeassistedmathematical} is based on similar intuition but instead focuses on code-aided mathematical reasoning tasks with the critic model iteratively. \citet{li2025startselftaughtreasonertools} instead relies on the external toolbox, such as code interpreters, to verify the process.

\begin{table*}[!htbp]
    \centering
    \resizebox{0.98\textwidth}{!}{
    \begin{tabular}{lll}
    \toprule
        \rowcolor{gray!10}
        \textbf{Category} & \textbf{Approach} & \textbf{Approach Description} \\
    \midrule
        \multirow{12}{*}{\textbf{Outcome}}
        & Naive ORM~\citep{cobbe2021training} & Naively process to train solution-level and token-level verifiers on labeled-dataset \\
        & OVM~\citep{yu2024ovm} & Train a value model under outcome supervision for guided decoding \\
        & Heuristic~\citep{deepseek-r1} & Heuristic check for domain-specific problems \\
        & Functional~\citep{lee2025evolvingdeeperllmthinking} & Functional scoring for task-specific problems \\
        & Bandit~\citep{sui2025metareasonerdynamicguidanceoptimized} & Train a bandit algorithm to learn how to verify \\
        & Generative Verifier~\citep{zhang2025generativeverifiersrewardmodeling} & Exploit the generative ability of LLM-based verifiers via reformulating the verification \\
        & Self-Reflection Feedback~\citep{li2025learningreasonfeedbacktesttime} & formulate the feedback utilization as an optimization problem and solve during test-time \\
        & Discriminator~\citep{chen2024tree} & SFT a domain-specific LM as a discriminator \\
        & Unit Test~\citep{saadfalcon2024archonarchitecturesearchframework} & Verify each sample as unit tests \\
        & XoT~\citep{liu2023plan} & Passive verification from external tools and Activate verification via re-thinking \\
        & WoT~\citep{zhang2024wrongofthoughtintegratedreasoningframework} & Multi-Perspective Verification on three aspects: Assertion, Process, and Result \\
        & Multi-Agent Verifiers~\citep{lifshitz2025multiagent} & Multi-Perspective Verification without explicit semantic meanings \\
    \midrule
        \rowcolor{gray!10}
        & Naive PRM~\citep{lightman2023let} & SFT an LM as a PRM on each reasoning step over mathematical tasks \\
        \rowcolor{gray!10}
        & State Verifier~\citep{yao2023tree} & SFT an LM as a state verifier and evaluate states either independently or jointly \\
        \rowcolor{gray!10}
        & Deductive PRM~\citep{ling2023deductive} & Deductively verify a few statements in the process \\ 
        \rowcolor{gray!10}
        & Self-Evaluation~\citep{xie2023selfevaluation} & Prompting the same LM to evaluate the current step given previous ones \\
        \rowcolor{gray!10}
        & Lego-prover~\citep{wang2023lego} & Evaluate lemma in Theorem Proving \\
        \rowcolor{gray!10}
        & PoT~\citep{chen2023program} & delegate computation steps to an external language interpreter \\
        \rowcolor{gray!10}
        & Tool~\citep{li2025startselftaughtreasonertools} & Relies on external toolbox for verification \\
        \rowcolor{gray!10}\multirow{-8}{*}{\textbf{Process}}
        & V-STaR~\citep{hosseini2024vstartrainingverifiersselftaught} & Verifier trained on both accurate and inaccurate self-generated data \\
    \bottomrule
    \end{tabular}}
    \caption{Summary of Certain Verification Techniques.}
    \label{tab:verification}
\end{table*}

\subsubsection{Search}
\label{subsec:search}

Search is also a frequently used component during the test-time scaling. LLMs pre-trained on vast amounts of online data, can be viewed as a compression of real-world knowledge. However, standard inference tends to underutilize their capacity. Search, being a classic yet working technique in retrieving relevant information from vast databases, can be utilized to fully exploit the capability of LLMs by exploring their potential options in a structured manner. Existing test-time scaling approaches based on search techniques demonstrate significant performance increases over complex tasks, such as complex mathematics, etc.

\citet{yao2023tree} explores the potential of search by decomposing the output samples into multiple thoughts and organizing them in a tree structure. Based on only Naive tree search algorithms, such as depth-first search and breath-first search, it demonstrates superior performance on reasoning tasks. 
Monte-Carlo Tree Search~\citep{coulom2006efficient}, being a classical and powerful search algorithm, also shines its light on better exploiting the hidden knowledge of LLMs.
\citet{chaffin2022ppl} adopts MCTS during the decoding stage guided by discriminators for constrained textual generation.
\citet{zhang2023planning} further extends the MCTS to enhance the planning ability in code generation via looking ahead.
\citet{tian2024toward} incorporates the MCTS as a critical component in the self-improving framework for LLM.
\citet{wan2024alphazero} tailors the search algorithm to tackle problems requiring long-horizon planning and deep tree structure for searching.
\citet{chen2024tree} further identifies that discriminators are the key bottleneck in search-enhanced planning.
\citet{gandhi2024streams} systematizes the search process in a unified language and proposes to train an LLM with data and feedback from the search process.
\citet{wu2024scaling} empirically analyzes various search algorithms and designs a reward-balanced search algorithm toward Pareto-optimal test-time scaling.
\citet{beenching2024scaling} further extends the beam search by incorporating diversity consideration.

Apart from searching within the tree structure, \citet{Besta2024graph} models the output as a graph search problem.
\citet{xie2023selfevaluation} proposes a stochastic beam search solution based on self-evaluation for reasoning tasks.
\citet{pan2025coatchainofassociatedthoughtsframeworkenhancing} enhances MCTS with proposed associative memory to dynamically update its knowledge base.
\citet{li2025reasoningaslogicunitsscalingtesttimereasoning} proposes to solve the reasoning process as constructing a control flow graph with each node indicating a logic unit.


\subsubsection{Aggregation}
\label{subsec:aggregation}
Aggregation techniques consolidate multiple solutions into a final decision to enhance the reliability and robustness of model predictions at test time. 
Based on how the final output is generated, we empirically categorize them into two key classes: (i) Selection, which selects the best-performed sample among all candidates, where the selection criteria may vary across different approaches; and (ii) Fusion, which fuse multiple samples into one through tricks like weighting or generation.

\paragraph{Selection} 
In this category, the aggregation process can be viewed as a selection problem. 
One well-known example is to select the most consistent answer, commonly known as \textit{self-consistency}. \citet{wang2023selfconsistency} improves accuracy by leveraging statistical redundancy—if different reasoning paths converge to the same conclusion, the answer is more likely to be correct. Self-consistency effectively reduces variance in model outputs and mitigates occasional hallucinations. However, as the final output is voted based on consistency, inaccurate and low-quality samples would inevitably influence the output quality. Therefore, various approaches are proposed to filter the candidates before voting. \citet{chen2024are} incorporates an LM as a filter, while \citet{wu2025lessunderstandingchainofthoughtlength} proposes a Length-filtered vote, where prediction uncertainty is adopted as a proxy to filter reliable CoT length. 

Best-of-N~\citep{irvine2023rewarding} follows the same process but replaces the self-consistency criteria with scalar scores generated by external verifiers. 
\citet{song2024good} further demonstrates that best-of-N on small LLMs can yield competitive performance against SOTA propriety models.
\citet{munkhbat2025selftrainingelicitsconcisereasoning} attaches a few-conditioning filtering before the best-of-N selection. This aims to alleviate its sample inefficiency and achieves significant length reduction.
Motivated by particle filtering, \citet{puri2025probabilisticinferenceapproachinferencetime} proposes to consider filtering upon the samples.
\citet{sessa2024bondaligningllmsbestofn} went one step further in reducing sample inefficiency. It tunes the best-of-N results into the LM via RLHF.
With the blooming of the agentic approach, \citet{parmar2025plangenmultiagentframeworkgenerating} proposes a selection agent considering complex factors with both historical and current status.
Apart from selecting samples from one single LM, \citet{ong2025routellm} views the selection of samples generated by weak and strong LLMs as a routing problem and proposes constraints on computation costs. 

\paragraph{Fusion}
Directly selecting the final output sample among candidates may yield unsatisfactory results, especially when the sample quality of candidates is low. Fusion approaches propose to merge multiple samples into one to solve such a problem.
\citet{brown2024large} and \citet{li2023making} extend the idea from Best-of-N and weigh each sample by its score from external verifiers.
\citet{jiang2023llm}, on the other hand, directly prompts another LLM as a summarizer to merge multiple selected samples.
\citet{li2025llmsgeneratebetteranswer} shares similar intuition by replacing the majority voting in self-consistency~\citep{wang-etal-2024-math} with generative self-aggregation.
\citet{li2025reasoningaslogicunitsscalingtesttimereasoning} also adopts LLM as the synthesizer, given the intermediate consideration in previous steps.

\begin{table}[!htbp]
    \centering
    \resizebox{0.98\textwidth}{!}{
    \begin{tabular}{l|l|c|l|l}
    \hline
        \rowcolor{gray!10}
        Category & Approach & External Verifier & Approach Description & Also Utilized in \\
    \hline
        & Majority Voting~\citep{wang2023selfconsistency} & \xmark & Select the most common sample & \cite{chen2024are} \\
        & Best-of-N~\citep{irvine2023rewarding} & \cmark & Select the highest scored sample & \cite{song2024good} \\
        & Few-shot BoN~\citep{munkhbat2025selftrainingelicitsconcisereasoning} & \cmark & BoN with few-shot conditioning \\
        \multirow{-4}{*}{Selection}
        & Agentic~\citep{parmar2025plangenmultiagentframeworkgenerating} & \xmark & agent considering both current and previous status \\
    \hline
        \rowcolor{gray!10}
        & Weighted BoN~\citep{li2023making} & \cmark & Weight each sample by its score & \cite{brown2024large} \\
        \rowcolor{gray!10} 
        & Synthesize~\citep{jiang2023llm} & \xmark & Fuse the selected samples via GenAI & \cite{wang2025mixtureofagents,li2025reasoningaslogicunitsscalingtesttimereasoning} \\
        \rowcolor{gray!10} \multirow{-3}{*}{Fusion}
        & Ensemble Fusion~\citep{saadfalcon2024archonarchitecturesearchframework}& \xmark & Conduct ensemble before fusion & \\
    \hline
    \end{tabular}}
    \caption{Summary of Certain Aggregation Techniques. BoN stands for Best-of-N.}
    \label{tab:aggregation}
\end{table}



\section{Where to Scale} 
\label{sec:where2scale}

\TTS can substantially enhance LLMs' performance across diverse real-world scenarios. We systematically categorize these scenarios into representative domains, detailing the characteristic challenges, critical evaluation criteria, and representative benchmarks that illustrate the practical value of TTS. Here, we also list a brief summary of various benchmarks in Table~\ref{tab:benchmark-summary-1}.

\subsection{Reasoning-intensive Tasks}
\label{sec:reasoning}

Reasoning-intensive tasks require structured, explicit, multi-step reasoning, precision, and rigorous correctness verification. These tasks challenge LLMs' ability to systematically decompose problems, iteratively refine solutions, and verify intermediate reasoning steps.

\paragraph{Mathematical Reasoning}

Mathematical tasks involve complex computations, logical inference, and iterative verification. Key challenges for \TTS methods include generating accurate step-by-step solutions, effectively verifying intermediate steps, and handling intricate reasoning logic. Representative benchmarks include MiniF2F~\citep{zheng2021minif2f}, AIME 2024~\citep{aime25}, MATH-500~\citep{zhang2024rest}, AMC 2023~\citep{guan2025rstarmath}, PutnamBench~\citep{tsoukalas2024putnambenchevaluatingneuraltheoremprovers}, MUSTARD~\citep{huang2024Mustard} and OlympiadBench~\citep{he2024olympiadbench}. These datasets span advanced competition-level math problems, emphasizing precise and explicit reasoning skills.

\paragraph{Programming \& Code Generation}
Coding tasks demand syntactic accuracy, executable correctness, and iterative debugging. Challenges for \TTS methods lie in generating correct implementations, debugging code iteratively, and efficiently exploring multiple coding solutions. Representative datasets include Codeforces~\citep{codeforce}, SWE-bench~\citep{jimenez2024swebench}, and LiveCodeBench~\citep{jain2025livecodebench}, each providing expert-level coding challenges that require rigorous logical thinking and implementation accuracy.

\paragraph{Game Playing and Strategic Reasoning}
Strategic reasoning tasks involve adaptive planning, interactive decision-making, and complex multi-round reasoning. \TTS methods must efficiently perform iterative search, adaptive inference, and dynamic interactions. A representative benchmark is SysBench~\citep{aime25}, which evaluates models' strategic reasoning in interactive tasks.


\paragraph{Scientific Reasoning}
Scientific problems typically require multi-domain knowledge integration across physics, chemistry, biology, and other disciplines. \TTS methods must demonstrate broad knowledge synthesis, multi-step reasoning, and accurate factual verification. Notable benchmarks include GPQA Diamond~\citep{rein2024gpqa} and MR-Ben~\citep{zeng2024mrben}, focusing on advanced scientific reasoning and integrated domain knowledge. Domain specific benchmarks, such as TP-Bench~\citep{chung2025tpbench}, UGPhysics~\citep{xu2025UGphysics}, PHYSICS~\citep{feng2025physics}, have also investigated recently.

\paragraph{Medical Reasoning}
Medical tasks involve diagnostic decision-making, clinical reasoning, and precise medical knowledge. The key challenge for \TTS here is ensuring reliable, accurate reasoning that mimics medical professionals’ decision logic. Representative datasets include JAMA Clinical Challenge~\citep{chen2025benchmarkinglargelanguagemodels}, Medbullets~\citep{chen2025benchmarkinglargelanguagemodels}, and MedQA~\citep{jin2020diseasedoespatienthave}. These benchmarks critically assess reasoning LLMs’ capabilities in diagnosis, treatment planning, and medical decision accuracy.

\subsection{Agentic Tasks}
The scaling of artificial intelligence agents can be classified into three distinct categories: scaling through design choice, scaling for emergent behavior analysis, and scaling through environmental interaction. In addition, many benchmarks have been proposed for agentic task training, attempting to provide extensive environment feedback.

\paragraph{Scaling Agents as Design Choice}
The first category involves scaling multi-agent systems as a deliberate design choice to enhance performance. A central question in this domain concerns the scaling laws of collaborative agents: specifically, how the progressive addition of collaborative agents impacts overall system performance. \citep{li2024more} studies the scaling of multi-agent system with sampling and voting. The experiments are conducted by using various LLMs of different sizes on diverse datasets covering reasoning and generation. Performance can be improved by increasing the ensemble size across various tasks, including GSM8k, MMLU, and HumanEval. \citep{chen2024more} have examined the relationship between the number of LLM calls and system performance, revealing non-monotonic behavior patterns. This non-monotonicity stems from varying query difficulties within tasks: while increased LLM calls enhance performance on simpler queries, they may diminish performance on more complex ones. \citep{qian2024scaling} studies the scaling of multi-round discussion in multi-agent collaboration on MMLU, HumanEval, SRDD, and SRDD under regular typologies including chain, tree, and graphs and irregular typologies. They found that the performance mostly follow a logistic growth pattern as the process of scaling agents. \citep{meyerson2025position} asks a more theoretical question on how task decompositions and assign multiple agents with different subtasks can be optimal?  It provides an asymptotic analysis with LLM primitives about the efficiency of such decomposed systems, leading future opportunities for scaling them.

\paragraph{Scaling Agent for Emergent Social Ability}
The second category focuses on scaling multi-agent systems to study emergent behaviors in large-scale simulations, particularly in social science applications. \citep{zhang2025understanding} studies large-scale multi-agent simulation over 300 agents under information asymmetry, focusing on phenomena including the emergence of information cocoons, the evolution of information gaps, and the accumulation of social capital. \citep{piao2025agentsociety} generates social lives for over 10k agents, simulating their 5 million interactions both among agents and between agents and their environment. The simulation is used as a testbed for to study four key social issues: polarization, the spread of inflammatory messages, the effects of universal basic income policies, and the impact of external shocks such as hurricanes. To support such large scale experiments, \citep{panvery} developed a user-friendly multi-agent infrastructure for future large-scale multi-agent experiments. 

\paragraph{Scaling Environment Feedback}
The third category involves scaling the interaction between agents and their environment to obtain richer feedback. \citep{hilton2023scaling} find out that an agent's intrinsic performance scales as a power law relative to model size and environmental interactions. However, constructing extensive environments to scale feedback and interactions for agent training remains challenging. \citep{chen2025scaling} tried to address this issue by training a reward model based on synthetic data to enable the agent to do Monte-Carlo Search during planning. This process typically involves using LLM-based agents to collect action trajectory demonstrations, using another LLM for generating contrasting negative trajectories, and training customized reward models based on the synthetic data collected. \citep{yu2024exact} tries to combine test-time search and self-learning by Reflective Monte Carlo Tree Search for richer feedback to enhance AI agents' ability to explore decision space on the fly. 

\paragraph{Simulated Environment for Agentic Tasks}
Agentic tasks involve realistic and interactive environments, requiring complex planning, iterative reasoning, and effective tool utilization. \TTS methods face challenges such as optimal stepwise planning, adaptive decision-making, tool integration, and iterative refinement. Representative benchmarks include WebShop~\citep{yao2023webshop}, WebArena~\citep{zhou2023webarena}, SciWorld~\citep{wang2022sciworld}, and TextCraft~\citep{prasad2024adaptasneededdecompositionplanning}. These datasets provide realistic interactive scenarios, emphasizing iterative decision-making and effective tool usage. Recent advances in scaling LLM-driven autonomous agents center on improved planning, memory, and self-optimization techniques. For example, ARMAP~\citep{chen2025scalingautonomousagentsautomatic} automatically learns a reward model from unlabeled environment interactions to score and guide an agent's actions, circumventing the need for human-labeled feedback and improving multi-step decision-making.

\subsection{Others}
\label{sec:generalpurpose}
These tasks require broad, general-purpose reasoning capabilities, creativity, and subjective evaluation of outputs. 

\paragraph{General}
To achieve general objectives, many efforts have collected numerous official, public datasets that are challenging for humans but are not exclusive to any particular domain. Representative benchmarks include AGIEval~\citep{zhong2024agieval}, MMLU-Pro~\citep{wang2024measuring}, and Gaokao~\citep{guan2025rstarmath}. These benchmarks may cover multiple aspects of language models and aim to test their general performance.

\paragraph{Open-Ended Tasks}
\TTS methods must enhance output diversity, quality, and coherence, balancing creativity and correctness. Representative benchmarks include AlpacaEval2.0~\citep{dubois2024lengthcontrolledalpacaevalsimpleway}, ArenaHard~\citep{li2024crowdsourceddatahighqualitybenchmarks}, IF-Eval~\citep{zhou2023instructionfollowing}, and C-Eval~\citep{huang2023ceval}, which collectively evaluate subjective, open-ended, and general-purpose reasoning.

\paragraph{Knowledge-intensive Tasks}
Knowledge-intensive tasks require LLMs to retrieve and synthesize factual knowledge from external sources, ensuring accuracy and reducing hallucinations. \TTS challenges center around effective retrieval-augmented reasoning, iterative verification, and multi-source aggregation. Representative benchmarks include SimpleQA~\citep{wei2024measuringshortformfactualitylarge}, C-SimpleQA~\citep{he2024chinesesimpleqachinesefactuality}, and FRAMES~\citep{krishna2025factfetchreasonunified}, emphasizing factual correctness and retrieval-based reasoning.

\paragraph{Evaluation Tasks}

Evaluation tasks require LLMs to act as judges, also known as Generative Reward Models (GRMs), to conduct comprehensive and in-depth quality assessments of the candidate responses, thus ensuring reliable evaluation results. Representative benchmarks in this field include RewardBench~\citep{lambert2024rewardbenchevaluatingrewardmodels}, JudgeBench~\citep{tan2025judgebenchbenchmarkevaluatingllmbased}, RMBench~\citep{liu2024rmbenchbenchmarkingrewardmodels}, PPE~\citep{frick2024evaluaterewardmodelsrlhf}, and RMB~\citep{zhou2025rmbcomprehensivelybenchmarkingreward}. Recent research~\citep{kim2025scalingevaluationtimecomputereasoning} has demonstrated that \TTS effectively enhances the evaluation reasoning capabilities of LLMs. For instance, CCE~\citep{zhang2025crowd} scales the evaluation by comparing the candidate responses with other crowd-generated responses, enabling TTS evaluation effects. EvalPlan~\citep{saha2025learningplanreason} achieves deeper evaluation by first generating a specific evaluation plan tailored to the candidate responses. SPCT~\citep{liu2025inferencetimescalinggeneralistreward} goes a step further by employing RL to generate evaluation principles, further activating the \TTS potential. Additionally, JudgeLRM~\citep{chen2025judgelrmlargereasoningmodels} has validated that training using the R1 method can effectively enhance the performance of RMs, while MAV~\citep{lifshitz2025multiagent} introduces multiple aspect verifiers. Notably, improving evaluator accuracy in Out-of-Distribution scenarios remains a critical issue, like Reward Hacking~\citep{skalse2025definingcharacterizingrewardhacking,shen2025exploringdatascalingtrends}, worthy of deeper exploration.

\paragraph{Multimodal Tasks}
Multimodal reasoning tasks demand effective cross-modal integration, iterative reasoning between modalities, and robust verification across visual and textual inputs. \TTS methods face challenges in modality fusion, iterative multimodal reasoning, and handling ambiguity across modalities. Representative benchmarks include MMMU~\citep{yue2024mmmu}, MathVista~\citep{lu2024mathvista}, MathVision~\citep{wang2024measuring}, CMMaTH~\citep{li2025cmmath}, and PGPS9K~\citep{zhang2023multimodalneuralgeometricsolver}, each testing multimodal reasoning across visual and textual modalities.

\begin{table*}[!ht]
\begin{adjustbox}{width=\textwidth,keepaspectratio}
\begin{tabular}{lccccc}
\toprule
\textbf{Benchmark} & \textbf{Size} & \textbf{Evaluation Criteria} & \textbf{Example Task} & \textbf{Key Features} & \textbf{Type} \\
\midrule
\multicolumn{6}{c}{\textbf{Reasoning-intensive Tasks}} \\
\midrule
FrontierMath~\citep{glazer2024frontiermath} & Hundreds & Exact match & Algebraic geometry & High complexity & \multirow{11}{*}{Math} \\
MATH~\citep{cobbe2021training} & 12.5K & Exact match & AMC/AIME-style & Structured reasoning &  \\
MiniF2F~\citep{zheng2021minif2f} & 1794 & Formal & Mixed & Formal Benchmark & \\
PutnamBench~\citep{tsoukalas2024putnambenchevaluatingneuraltheoremprovers} & 1709 & num-solved & Competition-level & Theorem-proving \\
MUSTARD~\citep{huang2024Mustard} & Varied & Pass@1 & Mixed & Auto-Generated \\
NuminaMath~\citep{numina_math_datasets} & 860K & Exact match, CoT & Olympiad-level math & Annotated reasoning &  \\
OmniMath~\citep{gao2025omnimath} & 4.4K & Accuracy & Math Olympiads & Advanced reasoning &  \\
GSM8K~\citep{zhang2024rest} & 8.5K & Accuracy & Grade-school math & Natural-language solutions &  \\
rStar-Math~\citep{guan2025rstarmath} & 747K & Pass@1 accuracy & Competition math & Iterative refinement &  \\
ReST-MCTS~\citep{zhang2024rest} & Varied & Accuracy & Multi-step reasoning & Reward-guided search &  \\
s1~\citep{muennighoff2025s1} & 1K & Accuracy & Math/science tasks & Controlled compute &  \\
\midrule
USACO~\citep{shi2024can} & 307 & Pass@1 & Olympiad coding & Creative algorithms & \multirow{4}{*}{Code} \\
AlphaCode~\citep{Li2022competition} & Thousands & Solve rate & Competitive coding & Complex algorithms &  \\
LiveCodeBench~\citep{jain2025livecodebench} & 511 & Pass@1 & Real-time coding & Live evaluation &  \\
SWE-bench~\citep{jimenez2024swebench} & 2.3K & Resolution rate & GitHub issues & Multi-file edits &  \\
\midrule
GPQA~\citep{rein2024gpqa} & 448 & Accuracy & Graduate STEM & Domain expertise & \multirow{9}{*}{Science} \\
PHYSICS~\citep{feng2025physics} & 1297 & Accuracy & University-level & Physics Problem \\
OlympicArena~\citep{huang2024olympicarena} & 11.1K & Accuracy & Multidisciplinary tasks & Multimodal reasoning &  \\
OlympiadBench~\citep{he2024olympiadbench} & 8.4K & Accuracy & Math/Physics Olympiads & Expert multimodal tasks &  \\
TheoremQA~\citep{chen2023theoremqa} & 800 & Accuracy & Theorem-based STEM & Theoretical application &  \\
TP-Bench~\citep{chung2025tpbench} & 57 & Grades & Mixed & Theoretical Physics \\
UGPhysics~\citep{xu2025UGphysics} & 5520 & Accuracy & University-level & Bilingual Physics Problem \\
PhysReason~\citep{zhang2025physreason} & 1200 & Step \& Answer & Mixed & Physics-based Reasoning \\
PHYBench~\citep{qiu2025phybench} & 500 & Accuracy \& EDD & Mixed & meticulously curated \\
\midrule
MedQA~\citep{jin2020diseasedoespatienthave} & 1.3K & Accuracy & Clinical diagnostics & Medical accuracy & \multirow{3}{*}{Medical} \\
JAMA ~\citep{chen2025benchmarkinglargelanguagemodels} & 1,524 & Accuracy & Challenging clinical cases & Expert-written explanations &  \\
Medbullets~\citep{chen2025benchmarkinglargelanguagemodels} & 308 & Accuracy & USMLE Step 2/3 clinical questions & Expert-written explanations &  \\
\midrule
\multicolumn{6}{c}{\textbf{Others}} \\
\midrule
AGIEval~\citep{zhong2024agieval} & 8K & Accuracy & College exams & Human-centric reasoning & \multirow{7}{*}{Basic} \\
MMLU-Pro~\citep{wang2024mmlupro} & 12K & Accuracy & Multidisciplinary tests & Deep reasoning complexity &  \\
C-Eval~\citep{huang2023ceval} & 13.9K & Accuracy & Chinese exams & Multidisciplinary reasoning &  \\
Gaokao~\citep{gaokao} & Varied & Accuracy & Chinese college exams & Broad knowledge & \\
Kaoyan~\citep{kaoyan} & Varied & Accuracy & Graduate entry exams & Specialized knowledge &  \\
CMMLU~\citep{li2024cmmlumeasuringmassivemultitask} & Varied & Accuracy & Multi-task Chinese eval & Comprehensive coverage &  \\
LongBench~\citep{bai2024longbenchbilingualmultitaskbenchmark} & Varied & Accuracy & Bilingual multi-task eval & Long-form reasoning &  \\
\midrule
IF-Eval~\citep{zhou2023instructionfollowing} & 541 & Accuracy & Instruction adherence & Objective evaluation & \multirow{4}{*}{Open-ended} \\
ArenaHard~\citep{li2024crowdsourceddatahighqualitybenchmarks} & 500 & Human preference & Open-ended creativity & Human alignment &  \\
Chatbot Arena~\citep{NEURIPS2023_91f18a12} & Varied & Human alignment & Chatbot quality & User-aligned responses &  \\
AlpacaEval2.0~\citep{dubois2024lengthcontrolledalpacaevalsimpleway} & 805 & Win rate & Chatbot responses & Debiased evaluation &  \\
\midrule

WebShop~\citep{yao2023webshop} & 1.18M & Task success & Online shopping & Real-world interaction & \multirow{4}{*}{Agentic} \\
WebArena~\citep{zhou2023webarena} & Varied & Task completion & Web navigation tasks & Adaptive decision-making &  \\
SciWorld~\citep{wang2022sciworld} & 30 tasks & Task-specific scores & Scientific experiments & Interactive simulation &  \\
TextCraft~\citep{prasad2024adaptasneededdecompositionplanning} & Varied & Success rate & Task decomposition & Iterative planning &  \\
\midrule
SimpleQA~\citep{wei2024measuringshortformfactualitylarge} & 4.3K & Accuracy & Short queries & Factual correctness & \multirow{3}{*}{Knowledge} \\
C-SimpleQA~\citep{he2024chinesesimpleqachinesefactuality} & 3K & Accuracy & Chinese queries & Cultural relevance &  \\
FRAMES~\citep{krishna2025factfetchreasonunified} & 824 & Accuracy & Multi-hop queries & Source aggregation &  \\
\midrule
RewardBench~\citep{lambert2024rewardbenchevaluatingrewardmodels} & 2,985 & Accuracy & Chat,Safety,Reasoning & Multiple Domains General Reward & \multirow{5}{*}{Evaluation} \\
JudgeBench~\citep{tan2025judgebenchbenchmarkevaluatingllmbased} & 350 & Accuracy & knowledge, reasoning, math, and coding & Challenging Tasks &  \\
RMBench~\citep{liu2024improvingmultistepreasoningabilities} & 1,327 & Accuracy & Visual math problems & subtle differences and style biases &  \\
PPE~\citep{frick2024evaluaterewardmodelsrlhf} & 16,038 & Accuracy & Instruction, Math, Coding, etc. & Real-world preference & \\ 
RMB~\citep{zhou2025rmbcomprehensivelybenchmarkingreward} & 3,197 & Accuracy & 49 fine-grained real-world scenarios & Closely related to alignment
objectives &  \\
\midrule
MMMU~\citep{yue2024mmmu} & 11.5K & Accuracy & Multimodal expert tasks & Multidisciplinary integration & \multirow{9}{*}{Multimodal} \\
MathVista~\citep{lu2024mathvista} & 6.1K & Accuracy & Visual math reasoning & Visual-math integration &  \\
MATH-Vision~\citep{wang2024measuring} & 3K & Accuracy & Visual math problems & Multimodal math reasoning &  \\
LLAVA-Wild~\citep{liu2023visual} & Varied & GPT-4 score & Visual QA & Complex visuals & \\
MM-Vet~\citep{yu2024mm} & Varied & GPT-4 evaluation & Integrated multimodal & Multi-capability eval &  \\
MMBench~\citep{liu2024mmbench} & 3.2K & Accuracy & Diverse multimodal & Fine-grained eval &  \\
CVBench~\citep{tong2024cambrian} & Varied & Accuracy & Vision tasks & High-quality eval &  \\
MMStar~\citep{chen2024we} & 1.5K & Accuracy & Vision-critical QA & Visual reliance &  \\
CHAIR~\citep{rohrbach2018object} & Varied & Hallucination rate & Image captioning & Object hallucination &  \\

\bottomrule
\end{tabular}
\end{adjustbox}
\caption{Summary of Benchmarks}
\label{tab:benchmark-summary-1}
\end{table*}

\section{How Well to Scale}
\label{sec:howwell2scale}

In this section, we classify the metrics used in evaluating the test-time scaling methods into four high-level dimensions: 
\textbf{Performance}, \textbf{Controllability}, \textbf{Scalability}, and \textbf{Efficiency}.
Each dimension captures an essential aspect critical to assessing test-time scaling approaches.

\subsection{Performance}
\label{subsec:performance}
Performance metrics assess the correctness of generated solutions.

\paragraph{Pass@1.} 
Pass@1 is one of the most widely used metrics for evaluating the correctness of a model’s first output attempt~\citep{deepseek-r1, li2025stesttimescaling, snell2024scaling, xie2025logic, kimi-k1.5, yang2025towards, yang2025reasonflux, hou2025advancing}. 
It measures the proportion of problems where the model’s first generated solution is correct. 
A correct solution means the one that exactly matches the ground-truth answer or passes all required validation checks, such as the exact answer match in mathematical benchmarks and private unit tests in coding tasks. 
Pass@1 is frequently used in tasks such as mathematical reasoning and coding benchmarks. 
In mathematical reasoning tasks such as \textbf{AIME 2024}~\citep{aime25} and \textbf{MATH-500}~\citep{zhang2024rest}, Pass@1 measures the percentage of exact matches between the model's answer and the ground truth.
In coding benchmarks such as \textbf{LiveCodeBench}~\citep{jain2025livecodebench} and \textbf{HumanEval-Mul}, Pass@1 evaluates the code correctness against hidden test cases. 

\paragraph{Pass@k (Coverage).} 
Pass@k extends Pass@1 by measuring whether at least one of the model’s $k$ sampled outputs is correct~\citep{brown2024large, snell2024scaling, li2025stesttimescaling}. Formally, Pass@k can be estimated using the unbiased estimator from~\citet{chen2021evaluating}:
\[
\text{Pass@k} = \frac{1}{n} \sum_{i=1}^{n} \left(1 - \frac{\binom{N - C_i}{k}}{\binom{N}{k}}\right),
\]
where $n$ is the number of problems, $N$ is the total number of samples per problem, and $C_i$ is the number of correct samples for the $i$-th problem. 
Pass@k is widely adopted in program synthesis and formal theorem-proving tasks, such as \textbf{CodeContests}~\citep{Li2022competition} and \textbf{SWE-bench Lite}~\citep{jimenez2024swebench}.

\paragraph{Cons@k (Consensus@k).}
Cons@k measures the majority vote correctness from $k$ independently sampled outputs~\citep{deepseek-r1, zeng2025revisiting}. Given $k$ responses generated by a model for a given problem, the majority-voted prediction is the most frequent answer. The answer is then compared against the ground truth. 
Cons@k is frequently used alongside pass@1 to assess the benefit of leveraging multiple samples. 
Larger values of $k$ (\emph{e.g.}, 16, 64) typically improve answer stability and accuracy but at the cost of increased compute.
This metric is especially valuable in tasks where single generations may be noisy or uncertain, and ensemble strategies can improve robustness. Cons@k has been widely adopted in mathematical reasoning benchmarks such as \textbf{AIME 2024}~\citep{aime25} and \textbf{MATH-500}~\citep{zhang2024rest}.

\paragraph{Arena-based Evaluation (Pairwise Win Rate).}
In addition to accuracy-oriented metrics, some studies adopt pairwise comparison metrics, where model outputs are compared against baselines using human or LLM-based judges~\citep{deepseek-r1, hou2025advancing}.
For instance, \textbf{LC-Winrate}~\citep{dubois2024lengthcontrolledalpacaevalsimpleway} adjusts win rates to control for response length, while \textbf{ArenaHard GPT-4 Judge}~\citep{li2024crowdsourceddatahighqualitybenchmarks} uses GPT-4-Turbo to score outputs from open-ended tasks.
These pairwise evaluation methods are especially common in generation tasks where qualitative assessments (\emph{e.g.}, fluency, coherence) matter.

\paragraph{Task-Specific Metrics.}
Certain domains employ specialized metrics. For example, \textbf{Codeforces} \textbf{Percentile} and \textbf{Elo Rating} are used to measure coding capabilities under competitive programming settings~\citep{deepseek-r1, kimi-k1.5}.
Percentile indicates how well a model performs relative to other participants, while Elo Rating reflects relative skill under tournament-based evaluations.

\subsection{Efficiency}
\label{subsec:efficiency}

Efficiency should not be treated as a monolithic concept. Instead, it spans multiple dimensions, encompassing both computational cost and the quality of the reasoning process. 
Different perspectives capture not only how many resources (e.g., tokens, FLOPs, memory) a model consumes, but also how effectively it reasons—whether it produces concise and goal-directed outputs, or falls into patterns of repetition and over-generation.

\subsubsection{Key Concern and Multi-Dimensional Perspectives of Efficiency}
\label{subsubsec:inefficiency_sources}

A central challenge in efficiency is the trade-off between the \emph{reasoning length} and \emph{solution quality}.  
The formal notion of \textbf{reasoning efficiency} captures this trade-off, as introduced by~\citep{qu2025survey}. 
Let $\mathcal{M}$ represent a language reasoning model evaluated over a distribution of tasks $\mathcal{T} \sim p(\mathcal{T})$, where each task consists of a dataset $\mathcal{D}$ and an associated question or objective. 
The reasoning efficiency $\eta(\mathcal{M})$ is defined as the expected ratio between solution quality $Q$ and computational cost $C$:
\begin{equation}
    \eta(\mathcal{M}) = \mathbb{E}_{\mathcal{T} \sim p(\mathcal{T})} \left[ \frac{Q(\mathcal{M}, \mathcal{D})}{C(\mathcal{M}, \mathcal{D})} \right],
\end{equation}
where $Q(\mathcal{M}, \mathcal{D})$ quantifies task performance (e.g., accuracy, exact match, or creativity), and $C(\mathcal{M}, \mathcal{D})$ measures the computation involved (e.g., number of generated tokens, FLOPs, or latency).
This concept highlights how excessively long reasoning chains may not yield proportionate gains in solution quality and often reflect redundancy in reasoning traces.

Recent work has identified multi-dimensional perspectives of inefficiency patterns in reasoning LLMs. 
These include \textbf{redundancy}, where models show excessive repetition of reasoning steps or thoughts, often restating questions or repeating points without contributing to solution quality~\citep{song2025prmbench, su2025token, luo2501o1}; \textbf{underthinking}, where models shift reasoning direction too early or fail to elaborate on promising ideas~\citep{wang2025thoughtsplaceunderthinkingo1like}; and \textbf{overthinking}, where models revisit or verify earlier steps excessively even for simple problems~\citep{luo2501o1, chiang2024over, chen2024not}. These behaviors contribute to higher computational cost without proportional gains in correctness, ultimately lowering overall reasoning efficiency.

\subsubsection{General Computational Cost Metrics}
\label{subsubsec:cost_metrics}

To quantify inference efficiency, we consider the following standard computational metrics:

\paragraph{Token Cost.}
Token cost measures the total number of tokens generated during inference, including intermediate reasoning steps and final outputs~\citep{welleck2024decoding, brown2024large, hou2025advancing, yang2025towards,xu2025softcotsoftchainofthoughtefficient,wang2025makepennycountdifficultyadaptive,aytes2025sketchofthoughtefficientllmreasoning}. This metric is especially important, as verbose reasoning typically leads to higher token consumption. Reducing token cost while maintaining performance is crucial for inference efficiency, particularly when operating under fixed computational budgets or API pricing constraints. 
In addition, inference efficiency metrics such as latency and throughput are critical in real-world applications, especially for high-throughput systems~\citep{welleck2024decoding}.

\paragraph{FLOPs-based Efficiency Analysis.}
FLOPs-based compute analysis has been widely adopted to quantify computational cost~\citep{kaplan2020scalinglawsneurallanguage, snell2024scaling, wu2024inference, teng2025atom}. Several recent works~\citep{snell2024scaling, wu2024inference} benchmark test-time scaling strategies, such as adaptive revisions and verifier-based search, against model scaling by plotting accuracy versus total inference FLOPs. This FLOPs-based evaluation can be used to determine whether test-time methods outperform larger models under equivalent compute budgets.

\paragraph{KV Cache Size.}
The \emph{KV cache size}~\citep{hooper2025etsefficienttreesearch} refers to the total memory footprint required to store the Key-Value cache across all trajectories and time steps during the test-time search process. As each unique generation path requires its own KV cache, methods with low KV sharing across trajectories tend to consume significantly more memory and incur higher latency.
By promoting KV cache sharing among trajectories, ETS reduces the total KV cache size, thereby improving throughput. For instance, ETS achieves up to $1.8\times$ KV cache reduction compared to REBASE, leading to $1.4\times$ faster inference on NVIDIA H100 GPUs, \emph{without compromising accuracy}.

\subsubsection{Reasoning Efficiency Metrics}
\label{subsubsec:underthinking}

Beyond standard compute metrics, reasoning inefficiency can also be measured through the following metrics.

\paragraph{Underthinking Score.}
The underthinking score~\citep{wang2025thoughtsplaceunderthinkingo1like}  quantifies cases where a model produces correct intermediate thoughts early in the reasoning chain but fails to reach a correct final answer.

Formally, the underthinking score $\xi_{\mathrm{UT}}$ is defined as:
\begin{equation}
\xi_{\mathrm{UT}} = \frac{1}{N_{\text{inc}}} \sum_{i=1}^{N_{\text{inc}}} \left(1 - \frac{\hat{T}_i}{T_i} \right),
\end{equation}
where $N_{\text{inc}}$ is the number of incorrect responses in the test set, $T_i$ is the total number of tokens in the $i$-th incorrect response, and $\hat{T}_i$ is the position (in tokens) of the first correct intermediate thought.
If no correct intermediate thought exists, $\hat{T}_i = T_i$, yielding a score of zero for that instance.
A higher $\xi_{\mathrm{UT}}$ indicates inefficient reasoning behavior where good ideas are introduced but not effectively pursued.

\paragraph{Outcome Efficiency.}
The outcome efficiency metric~\citep{chen2024not} quantifies how economically a model reaches the correct answer in multi-turn reasoning. It reflects whether additional solution rounds meaningfully contribute to accuracy. Empirically, the metric captures the ratio of efficient tokens—those used before the first correct answer appears—to the total number of output tokens. Formally, the outcome efficiency $\xi_{\mathrm{O}}$ is defined as:
\begin{equation}
\xi_{\mathrm{O}} = \frac{1}{N} \sum_{i=1}^{N} \sigma_i \cdot \frac{\hat{T}_i}{T_i},
\end{equation}
where $N$ is the number of test instances, $T_i$ denotes the total tokens generated for the $i$-th instance, and $\hat{T}_i$ is the number of tokens needed to reach the first correct solution. The correctness indicator $\sigma_i$ is set to 1 if at least one correct solution appears in the response, and 0 otherwise. A higher $\xi_{\mathrm{O}}$ suggests that the model typically finds correct answers early in the reasoning process, while a low value indicates inefficient use of computation due to overextended or redundant reasoning chains.

\paragraph{Process Efficiency.}
The process efficiency metric~\citep{chen2024not} evaluates how productively a model explores diverse reasoning strategies across multiple solution rounds. Rather than focusing solely on accuracy, it assesses the novelty of later solutions by measuring how many output tokens contribute to distinct reasoning approaches. Formally, the process efficiency $\xi_{\mathrm{P}}$ is defined as:
\begin{equation}
\xi_{\mathrm{P}} = \frac{1}{N} \sum_{i=1}^{N} \frac{D_i}{T_i},
\end{equation}
where $N$ is the number of test instances, $D_i$ is the number of tokens from non-redundant (i.e., distinct) solutions in the $i$-th response, and $T_i$ is the total number of tokens. A solution is considered distinct if its reasoning strategy differs significantly from those in earlier turns. 
A higher $\xi_{\mathrm{P}}$ value implies more efficient cognitive exploration, while a lower value indicates repetitive or self-verifying reasoning that fails to expand the model’s strategic diversity.






\subsection{Controllability}
\label{subsec:controllability}
Controllability metrics evaluate whether test-time methods can consistently adhere to pre-defined resource constraints such as compute budgets or output length targets.

\paragraph{Control Metric.} 
\citet{muennighoff2025s1} propose \textbf{Control} as a formal metric to quantify adherence to a specified compute budget range. It measures the fraction of test-time compute values that stay within given upper and lower bounds:
\[
\text{Control} = \frac{1}{|\mathcal{A}|} \sum_{a \in \mathcal{A}} \mathbb{I}(a_{\min} \leq a \leq a_{\max}),
\]
where $\mathcal{A}$ is the set of observed compute values such as thinking
tokens, and $\mathbb{I}(\cdot)$ is the indicator function. A score of 100\% denotes perfect adherence to the compute budget across all tasks.
Additionally, \citet{hou2025advancing} and \citet{yang2025towards} report experiments where models are evaluated under fixed token budgets, \emph{e.g.}, {1024, 2048, 4096}, to examine how well models meet pre-specified length or token constraints during reasoning.
Moreover, \citet{xie2025logic} and \citet{teng2025atom} impose explicit constraints on maximum output lengths to ensure test-time stability and prevent output truncation. 

\paragraph{Length Deviation Metrics.}
Mean Deviation from Target Length and RMSE of Length Deviation are introduced to quantify a model’s ability to control output length~\citep{aggarwal2025l1}:
\begin{itemize}[leftmargin=*]
    \item \textbf{Mean Deviation from Target Length} quantifies the average relative difference between the generated output length and the target length:
    \[
    \text{Mean Deviation} = \mathbb{E}_{x \sim D}\left[\frac{|n_{\text{generated}} - n_{\text{gold}}|}{n_{\text{gold}}}\right],
    \]
    where $n_{\text{generated}}$ is the model's output length and $n_{\text{gold}}$ is the target length.
    \item \textbf{Root Mean Squared Error (RMSE) of Length Deviation} captures the variance in length control:
    \[
    \text{RMSE} = \sqrt{\frac{1}{N} \sum_{i=1}^N \left(\frac{n_{\text{generated}, i} - n_{\text{gold}, i}}{n_{\text{gold}, i}}\right)^2}.
    \]
\end{itemize}
Lower values for both metrics indicate more stable and precise length control across samples.

\paragraph{\boldmath$k$--$\epsilon$ Controllability.} 
\citet{bhargava2024whatsmagicwordcontrol} propose \textbf{$k$--$\epsilon$ controllability} as a formal metric to characterize the prompt-based steerability of language models. Unlike metrics focused on compute or length constraints, this metric quantifies whether a model can be guided to produce a target output within a bounded prompt length and allowable deviation. Formally, a model is said to be $(k, \epsilon)$-controllable for a target output $y$ if there exists a prompt $p$ with $|p| \leq k$ such that the model outputs $y$ with probability at least $1 - \epsilon$:
\[
\Pr[\text{LLM}(p) = y] \geq 1 - \epsilon.
\]
By evaluating across different values of $k$ and $\epsilon$, one can map out the controllability landscape of a model. In practice, \citet{bhargava2024whatsmagicwordcontrol} measures this property on tasks such as next-token prediction in WikiText, finding that over 97\% of targets are reachable with a prompt of at most 10 tokens and an error tolerance $\epsilon \leq 0.05$. This metric provides a theoretical lens for quantifying how easily a model's outputs can be controlled via prompt design. While not directly tied to resource constraints, $k$--$\epsilon$ controllability offers valuable insight into the model's test-time responsiveness and has been used to compare inherent steerability across model families and sizes.

\subsection{Scalability}
\label{subsec:scalability}
Scalability metrics measure how effectively test-time scaling methods can leverage increased compute (e.g., token budgets, samples, inference steps) to improve performance.

\paragraph{Scaling Metric}
\citet{muennighoff2025s1} propose the \textbf{Scaling} metric, capturing the average slope of performance gains as compute increases:
\[
\text{Scaling} = \frac{1}{\binom{|\mathcal{A}|}{2}} \sum_{\substack{a, b \in \mathcal{A} \\ b > a}} \frac{f(b) - f(a)}{b - a}.
\]
This metric quantifies how effectively models improve accuracy or pass rates with additional computation.

\paragraph{Scaling Curves (Accuracy vs. Compute).}
Scaling curves are used to visualize how metrics such as accuracy, pass rate, or EM improve as token budgets, iteration depth, or the number of samples increase~\citep{aggarwal2025l1, teng2025atom, wu2024inference}. These plots help reveal diminishing returns and performance saturation at higher compute budgets.

\section{Organization and Trends in Test-time scaling}
\label{sec:organizationandtrends}

\begin{table*}[htbp]
    \centering
    \renewcommand{\arraystretch}{1.2} 
    \setlength{\tabcolsep}{8pt}
    \rowcolors{2}{gray!10}{white}
    \footnotesize
    \resizebox{\textwidth}{!}{
    \begin{tabular}{m{3.5cm}<{\raggedright} m{2cm}<{\centering} m{2cm}<{\centering} m{2cm}<{\centering} m{2.5cm}<{\centering} m{2.5cm}<{\centering} m{2.5cm}<{\centering} m{2.5cm}<{\centering} m{2.5cm}<{\centering} m{3.5cm}<{\centering}}
    \toprule
    \multirow{2}{*}{\textbf{Method}} & \multirow{2}{*}{\textbf{\textsc{What}}} & \multicolumn{6}{c}{\textbf{\textsc{How}}} & \multirow{2}{*}{\textbf{\textsc{Where}}} & \multirow{2}{*}{\textbf{\textsc{How Well}}}\\
    \cmidrule(r){3-8}
    \rowcolor{white}
    & & \textsc{SFT} & \textsc{RL} & \textsc{STIMULATION} & \textsc{SEARCH} & \textsc{VERIFICATION} & \textsc{AGGREGATION} & &\\
    \midrule
        {\makecell[l]{\textbf{DSC}\\\citep{snell2024scaling}}} & \makecell{Parallel, \\Sequential} & \xmark & \xmark & \xmark & \makecell{Beam Search,\\LookAhead Search} & Verifier & {\makecell{(Weighted) Best-of-N,\\Stepwise Aggregation}} & Math & {\makecell{Pass@1, FLOPs-\\Matched Evaluation}} \\
        {\makecell[l]{\textbf{MAV}\\\citep{lifshitz2025multiagent}}} & Parallel & \xmark & \xmark & Self-Repetition & \xmark & {\makecell[c]{\makecell{Multiple-Agent\\Verifiers}}} & Best-of-N & {\makecell{Math, Code,\\General}} & {\makecell[c]{BoN-MAV (Cons@k),\\ Pass@1}}\\ 
        {\makecell[l]{\textbf{Mind Evolution}\\\citep{lee2025evolvingdeeperllmthinking}}} & Sequential & \xmark & \xmark & Self-Refine & \xmark & Functional & \xmark & Open-Ended & {\makecell[c]{Success Rate, \\Token Cost}}\\
        {\makecell[l]{\textbf{Meta-Reasoner}\\\citep{sui2025metareasonerdynamicguidanceoptimized}}} & Sequential & \xmark & \xmark & CoT + Self-Repetition & \xmark & Bandit & \xmark & {\makecell[c]{Game,Sci,\\Math}} & {\makecell[c]{Accuracy, \\Token Cost}}\\
        {\makecell[l]{\textbf{START}\\\citep{li2025startselftaughtreasonertools}}} & \makecell{Parallel, \\Sequential} & \makecell{Rejection\\Sampling} & \xmark & Hint-infer & \xmark & Tool & \xmark & Math, Code & Pass@1\\
        {\makecell[l]{\textbf{AID}\\\citep{jin2025wellthinkingenhancingllm}}} & Sequential & \xmark & \xmark & {\makecell{Adaptive Injection\\Decoding}} & \xmark & \xmark & \xmark & {\makecell{Math, Logical,\\Commonsense}} & Accuracy \\
        {\makecell[l]{\textbf{CoD}\\\citep{xu2025chaindraftthinkingfaster}}} & Sequential & \xmark & \xmark & Chain-of-Draft & \xmark & \xmark & \xmark & \makecell{Math, Symbolic,\\Commonsense}  & {\makecell{Accuracy, Latency,\\Token Cost}} \\
    \midrule
        {\makecell[l]{\textbf{rStar-Math}\\\citep{guan2025rstarmath}}} & Hybrid & imitation & \xmark & \xmark & MCTS & PRM & \xmark & MATH & Pass@1\\
        {\makecell[l]{\citep{liu2025can}}} & \makecell{Parallel,\\Hybrid} & \xmark & \xmark & \xmark & {\makecell{DVTS,\\Beam Search}} & PRM & Best-of-N & Math & \makecell{Pass@1, Pass@k, \\Majority, FLOPS}\\
        {\makecell[l]{\textbf{Tree of Thoughts}\\\citep{yao2023tree}}} & Hybrid & \xmark & \xmark & {\makecell[c]{Propose prompt\\Self-Repetition}} & Tree Search & Self-Evaluate & \xmark & \makecell{GAME,\\Open-Ended} & \makecell{Success Rate, \\LLM-as-a-Judge} \\
        {\makecell[l]{\textbf{MindStar}\\\citep{kang2024mindstarenhancingmathreasoning}}} & Hybrid & \xmark & \xmark & \xmark & LevinTS & PRM & \xmark & MATH & \makecell{Accuracy, \\Token Cost} \\
        {\makecell[l]{\textbf{REBASE}\\\citep{wu2025inferencescalinglawsempirical}}} & Hybrid & \xmark & \xmark & \xmark & {\makecell{Reward Balanced\\Search}} & RM & \xmark & Math & \makecell{Test Error Rate,\\ FLOPs}\\
        {\makecell[l]{\textbf{RaLU}\\\citep{li2025reasoningaslogicunitsscalingtesttimereasoning}}} & Hybrid & \xmark & \xmark & Self-Refine & Control Flow Graph & Self-Evaluate & Prompt Synthesis & MATH, Code & Pass@1 \\
        {\makecell[l]{\textbf{PlanGen}\\\citep{parmar2025plangenmultiagentframeworkgenerating}}} & \makecell{Parallel,\\Hybrid} & \xmark & \xmark & MoA & \xmark & Verification agent & Selection Agent & \makecell{Math, General,\\Finance} & \makecell{Accuracy,\\F1 Score}\\
        {\makecell[l]{\citet{puri2025probabilisticinferenceapproachinferencetime}}} & Hybrid & \xmark & \xmark & \xmark & {\makecell{Particle-based\\Monte Carlo}} & PRM+SSM & Particle filtering & MATH & \makecell{Pass@1,\\Budget vs. Accuracy}\\
        {\makecell[l]{\textbf{Archon}\\\citep{saadfalcon2024archonarchitecturesearchframework}}} & Hybrid & \xmark & \xmark & \makecell{MoA,\\Self-Repetition} & \xmark & \makecell{Verification agent,\\Unit Testing} & \makecell{(Ensemble) Fusion} & \makecell{Math, Code,\\Open-Ended} & Pass@1, Win Rate\\
        {\makecell[l]{\textbf{AB-MCTS}\\\citep{misaki2025widerdeeperscalingllm}}} & Hybrid & \xmark & \xmark & \makecell{Mixture-of-Model} & AB-MCTS-(M,A) & \xmark & \xmark & Code & \makecell{Pass@1, RMSLE,\\ROC-AUC}\\
    \midrule
        \makecell[l]{\textbf{TPO}\\\citep{wu2024thinkingllmsgeneralinstruction}} & \makecell{Internal,\\Parallel} & \xmark & DPO & Think & \xmark & Judge models & \xmark & Open-Ended & Win Rate\\
        \makecell[l]{\textbf{SPHERE}\\\citep{singh2025selfevolvedpreferenceoptimizationenhancing}} & \makecell{Internal,\\Hybrid} & \xmark & DPO & Diversity Generation & MCTS & Self-Reflect & \xmark & Math & Pass@1\\
        \makecell[l]{\textbf{MA-LoT}\\\citep{wang2025malotmultiagentleanbasedlong}} & \makecell{Internal,\\Sequential} & imitation & \xmark & MoA & \xmark & Tool & \xmark & Math & Pass@k \\
        \makecell[l]{\textbf{OREO}\\\citep{wang2024offline}} & \makecell{Internal,\\Sequential} & \xmark & OREO & \xmark & Beam Search & Value Function & \xmark & Math, Agent & Pass@1, Success Rate \\
        \makecell[l]{\textbf{DeepSeek-R1}\\\citep{deepseek-r1}} & Internal & warmup & \makecell{GRPO,\\Rule-Based} & \xmark & \xmark & \xmark & \xmark & \makecell{Math, Code,\\Sci} & \makecell{Pass@1, cons@64,\\Percentile, Elo Rating,\\Win Rate} \\
        \makecell[l]{\textbf{s1}\\\citep{muennighoff2025s1}} & Internal & distillation & \xmark & Budget Forcing & \xmark & \xmark & \xmark & Math, Sci & \makecell{Pass@1, Control,\\Scaling}\\
        \makecell[l]{\textbf{o1-Replication}\\\citep{GAIR-o1p1}} & Internal & imitation & \xmark & \xmark & Journey Learning & PRM, Critique & Multi-Agents & Math & Accuracy\\
        \makecell[l]{\textbf{AFT}\\\citep{li2025draftsanswersunlockingllm}} & \makecell{Internal,\\Parallel} & imitation & \xmark & \xmark & \xmark & \xmark & Fusion & \makecell{Math,\\Open-Ended} & Win Rate \\
        \makecell[l]{\textbf{Meta-CoT}\\\citep{xiang20252reasoningllmslearning}} & \makecell{Internal,\\Hybrid} & imitation & meta-RL & Think & MCTS,A* & PRM & \xmark & \makecell{Math,\\Open-Ended} & Win Rate\\
        \makecell[l]{\textbf{ReasonFlux}\\\citep{yang2025reasonflux}} & \makecell{Internal,\\Sequential} & \xmark & \makecell{PPO,\\Trajectory} & Thought Template & Retrieve & \xmark & \xmark & Math & Pass@1\\
        \makecell[l]{\textbf{l1}\\\citep{aggarwal2025l1}} & Internal & \xmark & \makecell{GRPO,\\Length-Penalty} & \xmark & \xmark & \xmark & \xmark & Math & \makecell{Pass@1,\\Length Error}\\
        \makecell[l]{\textbf{Marco-o1}\\\citep{zhao2024marcoo1openreasoningmodels}} & {\makecell[c]{Internal,\\Hybrid}} & {\makecell{distillation,\\imitation}} & \xmark & Reflection Prompt & MCTS & Self-Critic & \xmark & Math & {\makecell{Pass@1, Pass@k}} \\
    \bottomrule
    \end{tabular}
    }
    \caption{Commonly-used combinations in existing literature when conducting inference scaling.}
    \label{tab:combination}
\end{table*}

\begin{figure}[!htbp]
    \centering
    \includegraphics[width=.98\linewidth]{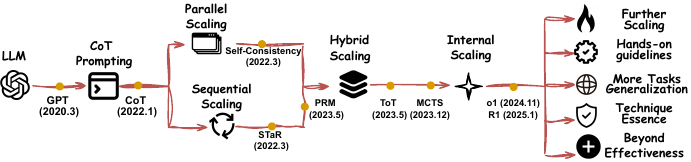}
    \caption{From Emergence to the Next Frontier, the Evolutionary Path of Test-Time Scaling.}
    \label{fig:timeline}
\end{figure}

Building on our taxonomy, we decompose the existing literature along multiple dimensions (Table~\ref{tab:combination}). As shown in Figure~\ref{fig:timeline}, these works, with different technical innovations, follow a broadly consistent path. From 2022 to 2023, researchers emphasized structured inference to guide LLMs in generating more complex solutions. In 2024, methods like PRM and MCTS enabled the automatic supervision of intricate reasoning trajectories, yielding richly annotated data for fine-tuning and improving \TTS performance. Subsequent approaches, such as o1 and R1, demonstrated that pure RL can also elicit comprehensive, logically sound reasoning. 

\begin{itemize}
    \item Crucially, these techniques are complementary rather than mutually exclusive: for instance, R1 necessitates an SFT-based warmup via rejection sampling. Therefore, achieving more powerful scaling requires systematically integrating these methods. Even within RL frameworks, practitioners should continue to leverage synthesized CoT approaches and incorporate structured inference strategies to tackle increasingly complex scenarios effectively.
    \item Researchers found that there does not exist one simple scaling solution that works for all problems. Increasingly, researchers tend to focus on optimal-scaling solutions~\citep{wu2024scaling,snell2024scaling}.
    \item The boundary between inference-based and tuning-based approaches is blurring. Consequentially, the target of scaling (\textit{what to scale}) changes between different stages. Certain papers, such as \citet{li2025draftsanswersunlockingllm, munkhbat2025selftrainingelicitsconcisereasoning}, tune the inference-based capability into the LLM by synthesizing high-quality data from inference-based approaches as the tuning data. Others, such as \citet{wan2024alphazero}, are proposing various techniques that better exploit the LLM's capability during both the training and inference stages.
\end{itemize}

\section{A Hand-on Guideline for Test-time Scaling}
\label{sec:handon}
In this section, we shift from theoretical categorizations to providing a practical, hands-on guideline for \textit{TTS}. Our goal is to offer clear, actionable instructions and technical pathways to facilitate effective SST deployment.

\begin{QABox}[Common Problems]

\Q{What kind of task does \TTS help?}

\A{Almost any task! While traditional reasoning tasks—such as Olympiad-level mathematics, complex coding, and game-based challenges—have been shown to significantly improve with \TTS, community observations suggest that \TTS can also enhance performance in open-ended tasks, such as comment generation or evaluation. However, due to the long-form nature of outputs and the lack of centralized, objective benchmarks, these tasks are inherently more difficult to evaluate quantitatively, making it harder to draw conclusive claims.
Beyond that, more realistic, complex, and long-horizon scenarios, like medical reasoning and law, have also shown promising gains through \TTS strategies.}

\vspace{1em}

\Q{If I want to quickly implement a \TTS pipeline, what are the essential paths I should consider? How can beginners use \TTS at a minimal cost?
}

\A{
Broadly speaking, there are three essential technical pathways for test-time scaling: i) Deliberate reasoning procedure at inference time, ii) imitating complex reasoning trajectories, and iii) RL-based incentivization. 
If your goal is to get a quick sense of the potential upper bound that a strong \TTS can bring to your task at a minimum cost, you can directly utilize a model that has been trained with (iii). If you want to develop a \TTS baseline at a minimum cost, you can start with (i). Once (i) yields a result that meets expectations, you can apply (ii) to further verify and generalize the outcome.
}

\vspace{1em}

\Q{Are these pipelines mutually exclusive? How should I design a frontier-level \TTS strategy?
}

\A{
These pipelines are by no means mutually exclusive—they can be seamlessly integrated. For instance, R1 inherently necessitates SFT through rejection sampling as a preliminary warmup step. When employing RL, practitioners should continue leveraging synthesized CoT methods and introduce additional structured inference strategies to tackle increasingly complex scenarios effectively. 
}
\vspace{1em}

\Q{Can training-time choices like SFT and RL affect the effectiveness of \textit{TTS}?}

\A{Yes, training-time strategies often shape the ceiling of what \TTS methods can achieve at inference.  
For example, SFT can provide strong reasoning priors that improve the stability and quality of scaling strategies like self-consistency or verifier-based search.  
On the other hand, RL-based fine-tuning can explicitly incentivize concise and correct reasoning chains, reducing overthinking or incoherent outputs at test time.  
Both approaches can also be combined: SFT initializes the policy, while RL further sharpens it—this is the strategy behind systems like DeepSeek-R1.  
Alternatively, RL-trained models can be distilled back into SFT models to reduce inference-time costs without losing the benefits of RL optimization.
}
\vspace{1em}

\Q{How can we improve the efficiency of \textit{TTS} in multi-turn setups?}

\A{Improving \textit{TTS} efficiency involves both structural design and learned behaviors:  
(1) \textit{Model-based}: Fine-tune models to generate more concise and purposeful reasoning traces. Reinforcement learning (RL) or reward-guided training can penalize overlong or redundant outputs while preserving correctness. Distillation from overthinking-prone models into lighter, more efficient ones can further reduce reasoning overhead.  
(2) \textit{Output-based}: Apply early stopping strategies based on verifier confidence or answer agreement to terminate reasoning once sufficient certainty is achieved. This avoids unnecessary continuation in well-understood problems.  
(3) \textit{Prompt-based}: Use prompt-level controls—such as token budgets, step limits, or task difficulty cues—to guide the model toward more targeted and efficient reasoning paths at inference.
These strategies together help reduce computational cost while maintaining strong solution quality during \textit{TTS}.
}
\vspace{1em}

\Q{What are some representative or widely-used \TTS methods that can serve as baselines?
}

\A{Parallel--Self-Consistency, Best-of-N; Sequential--STaR, Self-Refine, PRM; Hybrid--MCTS, ToT; Internal--Distilled-R1, R1.
}

\vspace{1em}

\Q{Is there an optimal go-to solution so far?
}

\A{No free lunch. Optimal computing is often dependent on the hardness and openness of the question.
}

\vspace{1em}

\Q{How should we evaluate the performance of a \TTS method? In addition to standard accuracy, what other aspects should we pay attention to?
}

\A{The evaluation is largely task-aware, but metrics like accuracy remain the most critical indicators. In addition, efficiency (the trade-off between performance and cost) is another key concern in practical settings. As \TTS becomes a more general-purpose strategy, researchers have also begun evaluating a range of secondary attributes, including robustness, safety, bias, and interpretability, to better understand the broader impacts of \TTS.
}

\vspace{1em}

\Q{Is there any difference when tuning other scaling formats into internal scaling, compared with directly using the original scaling format?
}

\A{Yes, one intuitive difference lies in the efficiency aspect. Internal scaling tends to yield higher efficiency as it only prompts the LM once, while other scaling techniques usually require multiple trials. However, internal scaling requires non-neglectable resources for tuning, making it less available for practitioners.}

\end{QABox}








\section{Challenges and Opportunities}
\label{sec:challenges}

\subsection{More Scaling is the Frontier}
\label{subsec:frontier}
Pushing AI toward more general intelligence, especially for complex tasks, test-time scaling has emerged as one of the most promising methodologies in the post-pretraining era. Given its transformative impact on reasoning-intensive tasks—as seen in models like OpenAI’s o1 and DeepSeek-R1—it is increasingly clear that realizing the full promise of test-time scaling remains a central pillar in advancing AGI. 
However, to push the frontier further, we need new and more effective strategies. There are some several promising research directions:

\paragraph{Parallel Scaling.} Parallel scaling improves solution reliability by generating multiple responses and selecting the best answer. Despite its effectiveness, parallel scaling remains has diminishing returns when coverage reaches saturation. A key challenge is how to enhance coverage, shifting from brute-force coverage expansion to a more guided, efficient process. Possible future advancements include: 
\begin{enumerate}
    \item \textit{Smart Coverage Expansion}: Instead of naive best-of-N sampling, a model could intelligently generate diverse reasoning paths, ensuring each sampled response explores a meaningfully different approach; 
    \item \textit{Verifier-Augmented Parallel Scaling}: Integrating real-time verification mechanisms could allow parallel samples to be filtered dynamically.
\end{enumerate}

\paragraph{Sequential Scaling.} Sequential scaling faces unique challenges, particularly in maintaining coherence and preventing error accumulation. A key issue is optimizing stepwise reasoning to avoid diminishing returns or reinforcing incorrect steps. Instead of naive iterative refinement, future advancements should focus on more adaptive and structured approaches to ensure each reasoning step meaningfully improves the final outcome. Possible directions include: 
\begin{enumerate}
    \item \textit{Structured Self-Refinement}: Rather than blindly refining the entire response, models could learn to target specific parts of their reasoning that require adjustment. 
    \item \textit{Verification-Enhanced Iterative Scaling}: Introducing real-time validation steps within the sequential reasoning process could prevent models from propagating early mistakes. This could involve running self-verification checks between iterations (\eg, checking consistency with known facts, comparing intermediate results to prior context, or re-computing specific logical steps). By selectively verifying before proceeding, models can ensure high-quality stepwise improvements instead of compounding errors.
\end{enumerate}
By addressing these challenges, sequential scaling can evolve beyond simple iterative refinement, becoming a highly adaptive, self-correcting reasoning paradigm that enables models to engage in goal-directed, long-horizon thinking.

\paragraph{Hybrid Scaling.} Hybrid scaling blends parallel and sequential methods, making it more adaptive and practical for real-world applications. Current test-time scaling methods are often highly specialized, limiting their generalizability.
To address these limitations, hybrid scaling can be improved in several ways:
\begin{enumerate}
    \item \textit{Generalized Hybrid Scaling Architectures}: research should focus on unifying test-time scaling mechanisms into a single framework that dynamically chooses the best strategy for different query types.
    \item \textit{Multi-Agent \& Interactive Scaling}: Expanding hybrid scaling beyond a single-agent reasoning process could allow multiple model instances to engage in structured debate, argumentation, or negotiation, improving solution reliability. While current hybrid scaling is mostly studied in controlled benchmarks, future work must consider its role in real-world applications.
\end{enumerate}

\paragraph{Internal Scaling.} Internal scaling allows on-the-fly computation modulation without external intervention. While this paradigm has demonstrated promising results, it also introduces unique challenges. 
\begin{enumerate}
    \item \textit{Effective Compute Allocation}: Ensuring that internal scaling allocates extra reasoning steps only where necessary is critical. If the model overthinks simple tasks or fails to extend reasoning on complex ones, the benefits of dynamic computation are lost.
    \item \textit{Stability and Consistency}: As models extend their own reasoning paths, they risk logical drift, hallucination, or over-complication. Unlike sequential scaling, which can incorporate external verification, internal scaling must maintain self-consistency without external guidance.
    \item \textit{Interpretability and Controllability}: Internal scaling happens implicitly, making it difficult to diagnose failures or regulate inference costs. Unlike parallel scaling (which provides multiple explicit outputs) or sequential scaling (which follows structured iterations), internal scaling lacks clear intermediate checkpoints, posing challenges for debugging and efficiency management.
\end{enumerate}

By addressing these challenges, internal scaling has the potential to maximize efficiency, enhance model adaptability, and push AI systems toward more autonomous, self-regulating reasoning.

\subsection{Clarifying the Essence of Techniques in Scaling is the Foundation}
\label{subsec:foundation}
While \textit{what to scale} continues to evolve and techniques further developing internally, such as PPO transitioning to GRPO, we observe that the core categories of scaling techniques remain relatively stable. For example, SFT and RL remain two of the most common approaches, though their roles and interactions have shifted over time. This raises an urgent need to deepen our understanding of how these fundamental techniques contribute to test-time scaling. 

Here, we raise some potential directions for further investigation:
\begin{enumerate}
    \item \textit{Theoretical Gaps in Scaling Techniques}: How do core techniques (SFT, RL, reward modeling) contribute to test-time scaling? how should SFT and RL be optimally combined?
    \item \textit{Re-evaluating Reward Modeling}: whether PRMs actually improve multi-step inference? Does the classic reward model incorporate noise and unnecessary complexity?
    \item \textit{Mathematical Properties of Test-Time Scaling}: How does performance scale with increased inference steps? Is there an optimal stopping criterion? Are there fundamental constraints on how much test-time scaling can improve reasoning performance?
    \item \textit{Chain-of-Thought Reasoning Priorities}: which aspects of chain-of-thought are most crucial for effective test-time scaling?
    \item \textit{Adaptive Test-Time Scaling}: How can we make a model automatically adjust its inference process based on the problem at hand? As empirical observations on certain property models~\citep{xai-gork3} show blindly scaling over test-time may lead to over-thinking.
    \item \textit{Thoughtology}: How do the reasoning patterns in its language help improve reasoning effectiveness by treating a finetuned reasoning model as an agent? Recent studies, such as \citet{marjanović2025deepseekr1thoughtologyletsthink,wu2024comparativestudyreasoningpatterns}, have also explored this question.
\end{enumerate}




\subsection{Optimizing Scaling is the Key}
\label{subsec:key}
As new \TTS methods proliferate, systematic evaluation and optimization become critical. We must comprehensively measure how different strategies perform regarding task accuracy and consider efficiency, robustness, bias, safety, interpretability, and more. Optimizing these aspects of \TTS is gradually emerging~\citep{zhang2025lightthinkerthinkingstepbystepcompression,huang2025efficienttesttimescalingselfcalibration} and will become an important part of future developments.

\subsection{Generalization across Domains is the Mainstream}
\label{subsec:mainstream}

We anticipate a wave of research extending test-time scaling into a wider range of domains, such as medicine and finance, where complex decision-making and structured reasoning are critical. This expansion is both inevitable and promising, as test-time scaling offers a powerful mechanism to enhance reasoning depth, adapt computation dynamically, and improve accuracy without requiring costly retraining. Beyond these fields, we can expect widespread applications in law, AI evaluation, open-domain QA, and other high-stakes or knowledge-intensive areas. Despite its potential, scaling test-time reasoning across domains presents several key challenges: 
\begin{enumerate}
    \item Balancing Cost and Accuracy: Unlike general NLP tasks, specialized domains often require strict computational efficiency and reliability; 
    \item Ensuring Domain-Specific Interpretability: In fields like medicine and law, outputs must be transparent and justifiable;
    \item Integrating External Knowledge \& Real-World Constraints: Many domains require retrieval-augmented generation, real-time data analysis, or interactive query refinement; 
    \item Future research must identify generalizable test-time scaling strategies that are robust across diverse reasoning tasks.
\end{enumerate}

By addressing these challenges, test-time scaling can become a foundational AI capability, enabling models to extend their own reasoning dynamically, adapt to real-world constraints, and generalize across specialized fields. This shift represents a paradigm change, where AI systems don’t just memorize knowledge—they actively scale their intelligence at inference to meet the demands of diverse, evolving tasks.

\section{Conclusion}
\label{sec:conclusion}

This is the first survey to decompose \TTS through a hierarchical taxonomy, offering a structured perspective that aids both conceptual understanding and the identification of individual contributions. Emphasizing practical utility, we introduce a hands-on guideline aligned with each taxonomy dimension, which we plan to expand over time. Based on this framework, we outline key trends, challenges, and opportunities shaping the future of \TTS research.

\section*{Author Contributions}
Below, we list the individual author contributions: Qiyuan Zhang and Fuyuan Lyu are core contributors who coordinate and finalize the full paper. Zexu Sun, Lei Wang, Weixu Zhang, Zhihan Guo, Wenyue Hua and Haolun Wu are significant contributors who are responsible for certain chapters of this paper. Yufei Wang provides the overall structures of the taxonomy and provides close supervision during the process. Niklas Muennighoff, Irwin King, Xue Liu, and Chen Ma provide insightful feedback and high-level suggestions on this survey overall.



\bibliographystyle{acl_natbib}
\bibliography{bib/dataset,bib/eval,bib/model_cards,bib/model_rep,general}

\appendix
\newpage
\begin{spacing}{0.2}
\tableofcontents
\end{spacing}

\newpage

\FloatBarrier

\section{Detailed Outcome Verification Methods}
\label{app:outcome_verification}

This appendix expands on the outcome verification techniques employed at test time in LLMs. Unlike training-time methods (\eg, RL fine-tuning), these techniques operate on the fly during inference, often by generating multiple solutions and using a \textit{proposer–verifier} framework.

\subsection{Verifier Model-Based Scoring}
The verifier, which is typically trained using human feedback or supervised data (\eg, as in~\cite{cobbe2021training,lambert2024rewardbenchevaluatingrewardmodels}), scores each candidate based on its expected correctness or quality. Variants include i) pairwise comparison verifiers~\citep{liu2025pairjudgermperformbestofn}, where candidates are compared against each other to determine a winner, ii) weighted voting systems~\citep{wettig2024quratingselectinghighqualitydata,li2024dnaevalenhancinglargelanguage} that use the verifier’s scores to combine outputs, iii) LLM-based verifiers that prompt LLM to perform evaluation instruction, like LLM-as-a-Judge~\citep{NEURIPS2023_91f18a12,zhang2025crowd,zhang2025reviseval}, LLM-based Evaluator~\citep{liu2023gevalnlgevaluationusing,xu2023instructscore,jiang2024tigerscorebuildingexplainablemetric}, Critic-based Model~\citep{gao2024llmcriticshelpcatch,mcaleese2024llmcriticshelpcatch}.

\subsection{Self-Consistency and Voting Mechanisms}
Self-consistency techniques generate multiple independent reasoning chains and choose the final answer based on majority voting~\cite{wang2023selfconsistency}. The underlying assumption is that if several chains converge on the same answer, that answer is more likely to be correct. Some approaches~\citep{taubenfeld2025confidenceimprovesselfconsistencyllms,mahmud2025enhancingllmcodegeneration} also incorporate confidence scores or soft-voting schemes to mitigate noise in individual outputs. In place of multiple samples from one model, one can also have multiple models~\citep{wan2025mammrefinerecipeimprovingfaithfulness,wu2025hiddenstrengthdisagreementunraveling,wang2025talkstructurallyacthierarchically,feng2024diverseagententropyquantifyingblackboxllm,chen2024reconcileroundtableconferenceimproves}: if a majority (or consensus) of these ``agents'' agree on an answer, trust it; if they diverge, it may trigger further scrutiny. This is effectively an ensemble vote. 

\subsection{Tool-Assisted and Heuristic Verification}
In domains like code generation or mathematical problem-solving, outcome verification can be implemented via direct execution or rule-based checks. For example, candidate programs are executed on sample test cases to ensure they produce correct results, while in math tasks, answers can be validated by plugging them back into the original equations. These approaches serve as an external check on the LLM's internal reasoning.

\paragraph{Execution-Based Verification.} In programming tasks, the ultimate test of correctness is running the code~\citep{tian2025codehaluinvestigatingcodehallucinations,ni2024nextteachinglargelanguage,yang2024exploringunleashingpowerlarge,ni2023leverlearningverifylanguagetocode}. For math problems, a simple heuristic is to verify the answer by plugging it back into the original equation or problem constraints. Similarly, if a puzzle answer must satisfy certain conditions, those can be programmatically checked.

\paragraph{Fact-Checking via Retrieval.} In open-domain QA or tasks that risk factual errors, search engines or knowledge bases serve as powerful verifiers~\citep{wei2024longformfactualitylargelanguage,vladika2024improvinghealthquestionanswering,asai2023selfraglearningretrievegenerate,peng2023checkfactstryagain}. An LLM may draft an answer, but then the system issues search queries (based on the answer’s claims) to find supporting evidence. If the retrieved documents contradict the LLM’s answer, the answer is likely incorrect and can be rejected or revised. Some frameworks generate answers in a ``closed-book'' fashion, then do a \textit{post-hoc} retrieval to validate facts. This idea overlaps with Retrieval-Augmented Generation~\citep{salemi2024searchenginemachinesunified}, but the focus is on post-generation validation – essentially checking if the answer aligns with external truth. 

\paragraph{Rule-Based Filters.} In some applications, simple heuristic filters~\citep{bai2022constitutionalaiharmlessnessai,NEURIPS2023_0764db11,weber2024redpajamaopendatasettraining} can automatically reject bad outputs. For a dialogue system, one might have a list of forbidden answers (certain unsafe or nonsensical replies) and if the model outputs one, the system can either regenerate or adjust it. These aren’t ``outcome-based'' in terms of correctness, but they verify the output against predefined rules of form and content.

\section{Representative Methods}

\subsection{Best-of-N}

The ``Best-of-N'' strategy is a \TTS approach in which a model generates $N$ candidate outputs for a given input and then selects the best one according to a chosen evaluation metric~\citep{wu2024scaling}. Mathematically, given an input $x$ and model $f$, one draws $N$ independent outputs $y_1,\dots,y_N \sim f(x)$ (\eg, via different random seeds or sampling strategies) and chooses the result $\hat{y} = \arg\max_{i=1}^N M(y_i)$, where $M$ is a quality scoring function. At the cost of additional inference compute, increasing $N$ raises the probability of obtaining a high-quality outcome (for example, if each attempt succeeds with probability $p$, then a best-of-$N$ run succeeds with probability $1 - (1-p)^N$). This technique leverages extra computation to boost performance~\citep{kang2025scalablebestofnselectionlarge} and has been applied in real-world settings ranging from complex reasoning and code generation with LLMs to enhancing image synthesis quality in diffusion models~\citep{ma2025inferencetimescalingdiffusionmodels}.

\subsection{Majority Voting}
Majority voting is a fundamental ensemble strategy for \TTS that aggregates multiple independent predictions to make a final decision. In this approach, each model or inference (voter) casts a vote for a predicted outcome, and the output chosen is the one with the highest number of votes (\ie, the mode of the predictions). Formally, given an ensemble of $M$ models $h_1, h_2, \dots, h_M$ each producing a prediction $h_m(x)$ for input $x$, the majority vote outcome is defined as 

\[
\hat{y} = \arg\max_{c} \sum_{m=1}^{M} \mathbf{1}\{\,h_m(x) = c\,\},
\]

where $\mathbf{1\{\cdot\}}$ is the indicator function and $c$ ranges over all possible classes or outputs. This test-time inference technique leverages additional computing at inference to improve reliability without retraining models, and it is widely used in real-world applications, such as combining votes of decision trees in a random forest, consolidating crowd-sourced annotations, or enhancing the consistency of answers from LLMs by selecting the most frequent response.

\subsection{Process Reward Model}
A Process Reward Model (PRM)~\citep{uesato2022solvingmathwordproblems,pfau2024lets} is a reward model designed to evaluate an entire reasoning trajectory on a step-by-step basis. Formally, given an input problem $x$ and a sequence of reasoning steps $z_1, z_2, \dots, z_T$ leading to a final output $y$, we can represent this full reasoning trace as:
\[
    S^T = (x, z_1, z_2, \dots, z_T, y),
\]
and define the PRM as a function that assigns a real-valued score:
\[
    r: S^T \to \mathbb{R},
\]
mapping a possible reasoning process $S^T$ to a reward score~\citep{choudhury2025processrewardmodelsllm,ma2025steplevelrewardmodelsrewarding}. Intuitively, $r(S^T)$ is higher when the reasoning process is logical, valid, and leads to a correct solution, and lower (or negative) when the reasoning is flawed. PRMs are typically trained on human or algorithmic annotations for each step, internalizing a notion of ``partial credit'' to evaluate correctness and relevance at each stage.

PRMs play a crucial role in \TTS strategies such as stepwise beam search and self-consistency verification. They have been successfully applied in mathematical reasoning, code generation, automated theorem proving, and decision-making tasks. By leveraging PRMs, models can optimize not only for correctness but also for process coherence, making AI systems more transparent and robust.

\subsection{MCTS}

Monte Carlo Tree Search (MCTS) is a simulation-based decision-making algorithm for sequential decision problems, often formalized as a Markov Decision Process (MDP). It incrementally builds a search tree by sampling many possible future trajectories (playouts) and using their outcomes to estimate the value of decisions. Unlike brute-force search, MCTS selectively explores the most promising actions by balancing exploration (trying unexplored or uncertain moves) and exploitation (favoring moves with high estimated reward). Each iteration of MCTS consists of four phases:

\begin{enumerate}
    \item Selection – Recursively select child actions that maximize a heuristic value until reaching a leaf node. A common selection strategy is the Upper Confidence Bound for Trees (UCT):
    \[
    \text{UCT}(a) = \frac{w_a}{n_a} + c\,\sqrt{\frac{\ln N}{n_a}},
    \]
    where $w_a$ is the total simulation reward, $n_a$ is the visit count for action $a$, $N$ is the total simulations from the parent state, and $c>0$ is an exploration constant.
    
    \item Expansion – Once a leaf state is reached, new child nodes are created by simulating unexplored actions.
    
    \item Simulation (Rollout) – Perform a Monte Carlo simulation by selecting actions to simulate a full episode to the end, providing an estimate of the node's value.
    
    \item Backpropagation – Propagate the simulation result back up the tree, updating the statistics of each node along the path.
\end{enumerate}

MCTS is well-suited for \TTS because its anytime nature allows flexible computation budgets. At test time, running MCTS for longer or with more rollouts leads to deeper search and better decisions. Notably, AlphaGo used MCTS at runtime to refine moves, significantly improving performance without additional training.

Researchers are leveraging MCTS to enhance test-time reasoning in other AI domains. MCTS-Judge improves code correctness evaluation by systematically exploring reasoning paths, raising verification accuracy significantly. Similarly, hybrid approaches integrate MCTS into generative model inference for problem-solving, such as solving Sudoku puzzles through sequential search.

By repeating these steps, MCTS concentrates simulations on the most promising branches. In the limit, MCTS value estimates converge to the optimal values in certain perfect-information games.

\subsection{Self-Refine}

Self-Refine~\citep{madaan2023selfrefine} is an advanced TTS technique that enables an LLM to iteratively improve its own outputs through self-generated feedback. Introduced by \citet{madaan2023selfrefine}, the Self-Refine framework is inspired by how humans revise a draft: The model first produces an initial answer, then critiques or evaluates that answer, and finally uses the critique to refine the answer. This feedback-refinement loop can be repeated multiple times, progressively polishing the output. Notably, Self-Refine requires no additional training data or fine-tuning – the same pre-trained model acts as the initial answer generator, the feedback provider, and the refiner.
For sufficiently powerful models, this self-iteration yields significantly better results, presumably because it is easier for a model to identify and fix errors in a given solution than to produce a perfect solution in one attempt. In essence, Self-Refine leverages test-time compute to let the model ``think twice (or more)'' about its answer, leading to higher-quality and more reliable outputs.

Formally, consider an input $x$ and a language model $M_\theta$ with parameters $\theta$, defining a conditional distribution $P_\theta(y \mid x)$ over possible outputs $y$. The Self-Refine procedure generates a sequence of outputs $y^{(0)}, y^{(1)}, \dots, y^{(T)}$ as follows:

\begin{enumerate}
    \item Initial Output Generation: The model first produces an initial response:
    \begin{equation}
        y^{(0)} = M_\theta(x).
    \end{equation}
    
    \item Feedback Generation: At each refinement step $t = 1,2,\dots,T$, the model evaluates the previous output and generates feedback:
    \begin{equation}
        f^{(t)} = M_\theta\big(x, y^{(t-1)}; \text{feedback-prompt}\big).
    \end{equation}
    
    \item Refinement Step: Using the generated feedback, the model updates its output:
    \begin{equation}
        y^{(t)} = M_\theta\big(x, y^{(t-1)}, f^{(t)}; \text{refine-prompt}\big).
    \end{equation}
\end{enumerate}

This feedback-refinement loop continues iteratively until a stopping condition is met, such as reaching a predefined number of iterations $T$ or detecting convergence in the output quality. The Self-Refine approach enhances model reliability by progressively improving its responses without requiring additional training.

\subsection{Tree-of-Thought}

Complex reasoning problems often require exploring different lines of thought before arriving at a correct solution. CoT prompting was a first step in this direction: CoT guides the model to produce a single sequence of intermediate reasoning steps (a linear chain) leading to the answer. This improves the model’s performance on tasks requiring multi-step logic by breaking the problem into a step-by-step narrative. However, CoT still follows a single path – if the model makes a wrong turn in the reasoning chain, it cannot recover because it doesn’t revisit earlier decisions. Tree-of-Thought~\citep{yao2023tree}, by contrast, generalizes CoT to a branching search. At each reasoning step, the model can generate multiple candidate thoughts instead of one, forming a tree of possibilities. It evaluates these candidates (using heuristics or self-evaluation prompts) and selects the most promising branch(es) to continue expanding~\citep{bi2024forest}. This test-time exploration allows the model to consider alternative approaches and scale up inference computation as needed – much like how a human might try different reasoning avenues for a hard problem. Researchers have categorized ToT and similar strategies (\eg, graph-of-thought) as "X-of-Thought" (XoT) reasoning methods, which significantly improve LLM reasoning by introducing iterative, structured inference without additional training.

ToT can be modeled as a search process through a state space of partial solutions, where each state encodes the sequence of thoughts (intermediate steps) explored so far. Let $S$ be the set of all possible reasoning states for a given problem. The initial state $s_0$ contains the problem statement, and a goal state $s \in S$ represents a complete solution.

\textbf{Thought Generation (State Transitions)}: At each step, the language model serves as a thought generator function $G$. Given the current state (context) $s$, the model generates a set of next-step thoughts:
\begin{equation}
    G(s) \rightarrow \{t_1, t_2, \dots, t_b\}
\end{equation}
where each $t_i$ represents a candidate next reasoning step. Each thought extends the current reasoning path, yielding a new state:
\begin{equation}
    s_i = s \oplus t_i
\end{equation}
where $\oplus$ denotes concatenation of the thought to the sequence.

\textbf{State Evaluation (Heuristic Function)}: To guide the search, ToT uses an evaluation function $f(s)$ that estimates the quality of a partial state $s$:
\begin{equation}
    f: S \rightarrow \mathbb{R}
\end{equation}
This function may be implemented by the model itself using a self-evaluation prompt or a scoring heuristic.

\textbf{Search Algorithm (Tree Expansion)}: ToT can employ different search strategies, including:
\begin{itemize}
    \item \textbf{Breadth-First Search (BFS)}: Expands all plausible thoughts at each depth, keeping the top $b$ best states based on $f(s)$.
    \item \textbf{Depth-First Search (DFS)}: Follows the most promising thought path deeply, backtracking if necessary.
\end{itemize}

Each strategy allows ToT to control computational budgets by limiting depth $d$ (number of steps) and branching factor $b$ (number of candidates per step).

\textbf{Solution Extraction}: A state $s$ is considered a valid solution if it satisfies the problem constraints. The search continues until:
\begin{enumerate}
    \item A goal state is reached.
    \item The computational budget (depth or number of states evaluated) is exhausted.
\end{enumerate}

This framework formalizes ToT as an organized search over the space of reasoning sequences, allowing models to iteratively refine and explore multiple potential solutions during test-time inference.

\subsection{Reinforcement Learning}

Reinforcement learning can play a pivotal role in unlocking effective \TTS for language models. The process of inference itself can be formulated as a sequential decision-making problem: at each step in generating a solution, \eg, each token in a reasoning chain or each attempt at an answer, the model (agent) must decide whether to continue reasoning, which direction to explore, or when to stop and output an answer. By training the model with RL, we can explicitly reward outcomes that lead to correct or high-quality answers, thereby encouraging the model to make better use of the extra inference steps available. This addresses a key challenge in \TTS: simply allowing a model to think longer doesn’t guarantee better answers unless the model knows how to productively use that extra time (it could otherwise repeat mistakes or terminate too early). RL provides a feedback-driven way to learn such behaviors. In fact, prior approaches to improve reasoning in LLMs often relied solely on imitation learning (learning from observed human or AI reasoning traces), which can limit a model to mimicking given patterns~\citep{hou2025advancing}. By contrast, RL enables self-exploration: the model can try diverse reasoning paths and learn from trial-and-error which strategies yield the highest reward (for example, reaching a correct solution). This means an RL-trained language model can learn dynamic inference policies—such as when to double-check an intermediate result or how to backtrack and correct itself if the reasoning seems to be going astray. Recent research indeed shows that combining chain-of-thought reasoning with reinforcement learning techniques leads to improved inference-time performance.

\end{document}